%% file: acl_latex.tex
\documentclass[11pt]{article}

\usepackage[final]{acl}

\usepackage{times}
\usepackage{latexsym}

\usepackage[T1]{fontenc}

\usepackage[utf8]{inputenc}

\usepackage{microtype}

\usepackage{inconsolata}

\usepackage{graphicx}
\usepackage{amsmath}
\usepackage{amssymb}
\usepackage{mathtools}
\usepackage{amsthm}
\usepackage[capitalize,noabbrev]{cleveref}

\usepackage[T1]{fontenc}
\usepackage{textcomp}
\usepackage{hyperref}
\usepackage{url}
\usepackage{tcolorbox}
\usepackage{multirow}
\usepackage{multicol}
\usepackage{hhline}
\usepackage{color}
\usepackage{wasysym}
\usepackage{pifont}
\usepackage{wrapfig}
\usepackage{colortbl}
\usepackage{threeparttable}
\usepackage{enumitem}
\usepackage{adjustbox}

\usepackage[utf8]{inputenc} 
\usepackage[T1]{fontenc}    
\usepackage{hyperref}       
\usepackage{url}            
\usepackage{booktabs}       
\usepackage{amsfonts}       
\usepackage{nicefrac}       
\usepackage{microtype}      
\usepackage{multirow}        
\usepackage{arydshln}        
\usepackage[table]{xcolor}   
\usepackage{algorithm}
\usepackage{enumitem}
\usepackage{xcolor}
\usepackage{tcolorbox}
\usepackage{fancyvrb}
\usepackage{pifont}
\usepackage{subcaption}
\usepackage{minted}
\usepackage{newunicodechar}
\usepackage{wrapfig}
\usepackage{stmaryrd}  
\usepackage{mathtools} 
\usepackage{fvextra}
\usepackage{float}
\usepackage{graphicx}
\usepackage{placeins}
\usepackage{adjustbox}
\def\ModelNameL{\textbf{SpatialMem}}
\def\DatasetName{\textbf{MetricBench}}

\newcommand{\second}{\cellcolor[HTML]{bed6f1}}
\newcommand{\best}{\cellcolor[HTML]{acc1e9}}
\definecolor{kleinblue}{rgb}{0,0.18,0.65}
\hypersetup{colorlinks=true,citecolor=kleinblue, linkcolor=black, urlcolor=kleinblue}

\title{SpatialMem: Building Quantitative Spatial Reasoning Agents with Hierarchical Memory}

\author{Zinan Zheng \\
  Hong Kong University of Science \\and Technology
(Guangzhou) \\
  \texttt{zzheng078@connect.hkust-gz.edu.cn} \\\And
  Nuo Chen \\
  HY Foundation Model\\ Team, Tencent \\\And
  Jia Li \\
  Hong Kong University of Science \\and Technology
(Guangzhou)\\
  \texttt{jiale@ust.hk} \\}

\begin{document}
\maketitle
\begin{abstract}
Vision Language Models (VLMs) have shown remarkable capabilities in qualitative spatial reasoning, but remain limited in quantitative spatial reasoning. Current work leverages the VLM agent to invoke external vision tools to perform computation. However, these methods usually bind each functional requirement to a single predefined vision specialist, implicitly resulting in fixed and rigid tool pipelines. In practice, addressing diverse spatial scenarios and tasks demands a broader diversity in pipeline orchestration and vision specialist selection to achieve robust spatial reasoning. However, such diversity leads to a combinatorial execution space, making it challenging for the agent to orchestrate the optimal pipelines and specialists for different scenarios.

In this work, we propose \ModelNameL, a memory framework that enables a VLM agent to accumulate tool-use experience and leverage that memory to orchestrate vision specialists to solve the spatial reasoning problem. \ModelNameL~ builds a two-level memory structure, with task memory capturing tool orchestration and scene memory capturing tool selection. It further introduces a dual-state trajectory optimization process to preserve successful tool experiences and refine suboptimal ones. Moreover, existing spatial reasoning benchmarks suffer from quality issues and limited coverage of scene types and tasks. To address these gap, we propose \DatasetName, a benchmark covering both indoor and outdoor scenes and three quantitative spatial reasoning tasks. Our method achieves state-of-the-art results on \DatasetName~, Q-Spatial, and SpatilRGPT-Bench, outperforming best VLM baselines by 32\% on \DatasetName. Additionally, the memory design yields a 9.1\% absolute gain over a memory-free agent on \DatasetName. Codes are available at ~\url{https://github.com/compasszzn/SpatialMem}.
\end{abstract}
\input{section/intro}

\input{section/related}
\input{section/formulation}
\input{section/method}
\input{section/experiment}
\input{section/conclusion}

\section*{Acknowledgements}
This research was supported by the Guangdong Provincial Talent Program under Grant (No.2024TQ08X366).
\section*{Limitations}
Our framework is designed specifically for quantitative spatial reasoning and primarily focuses on spatial question answering tasks. As a result, the current memory hierarchy relies on predefined scene and task categories, which restricts its immediate generalizability to non-spatial or purely textual long-horizon challenges, such as code generation or multi-step logic programming. In addition, the effectiveness of memory evolution depends on sufficient interaction experience and reliable evaluation feedback, which serve as key factors for stable memory updates.
\bibliography{custom}
\newpage
\appendix

\input{section/appendix}
\label{sec:appendix}

\end{document}

%% file: section/intro.tex
\section{Introduction}

Recent advances in Vision-Language Models (VLMs)~\cite{zhou2026unitybench,zheng2026weathersyn,chen2024breaking,zheng2026mesh,zhao2026channelmts} have demonstrated remarkable capabilities across a wide range of visual understanding and reasoning tasks. However, spatial reasoning remains a fundamental capability for VLMs, requiring them to understand object relationships and their surrounding environments. It is critical for embodied systems~\cite{li2025embodied},  robot navigation, and spatial question answering~\cite{ma2024llms}. Recent work~\cite{chen2024spatialvlm,chen2026sd,cheng2024spatialrgpt} achieves strong performance on qualitative tasks (e.g., relative position, distance, and size comparison) but struggles with quantitative reasoning (e.g., exact metric sizes or object distances estimation). These methods rely on large-scale image-text QA datasets, which often encourage models to exploit statistical biases and linguistic shortcuts rather than develop genuine geometric understanding. To address this, recent studies~\cite{han2025tiger,chen2025spacetools,marsili2025visual} instead treat VLMs as reasoning agents that perform multi-step reasoning, invoke external tools, and generate executable code for more precise spatial reasoning.
\begin{figure}[t]
\begin{center}
\centerline{\includegraphics[width=\columnwidth]{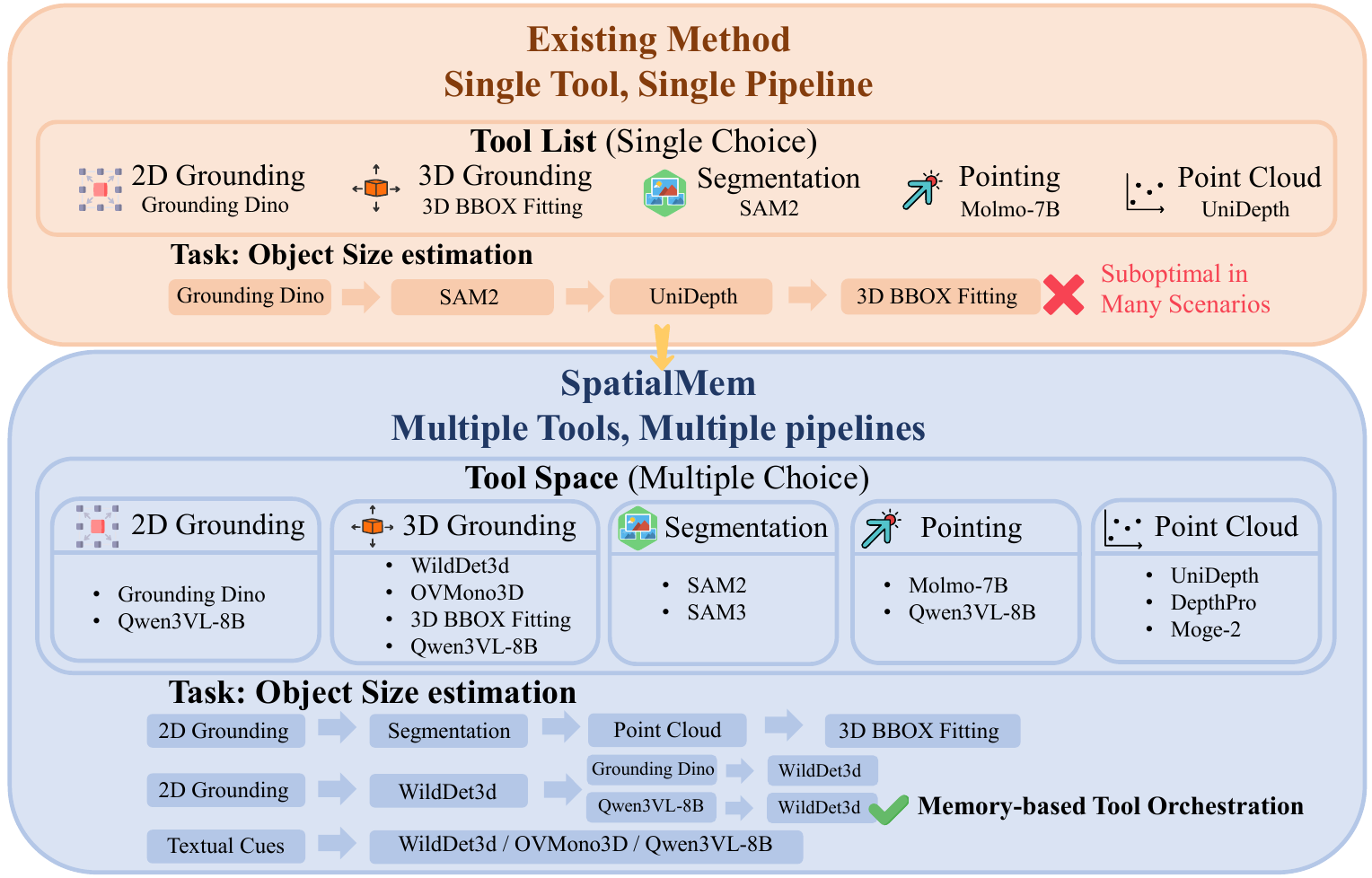}}
\caption{Different pipelines and specialists exhibit varying effectiveness due to biases induced by tools, as well as differences across scenes and tasks.}
\vspace{-5ex}
\label{fig:intro}
\end{center}
\end{figure}

Despite this progress, existing methods operate within a highly constrained paradigm. As shown in Figure~\ref{fig:intro}, they typically bind each functional requirement to a fixed vision specialist, implicitly forming a static tool pipeline. In practice, each functional requirement can be supported by multiple vision specialists, and the effectiveness of different pipelines and components varies substantially across geometric conditions, scene types, and task categories. This variability implies that relying on a single specialist severely limits adaptability and robustness. Therefore, it is crucial for an agent to accumulate tool-use experience and to discover optimal macro-level pipeline structures as well as micro-level specialist assignments tailored to specific scenarios.




To address this gap, we introduce \ModelNameL, a novel hierarchical memory framework that empowers Vision-Language Model (VLM) agents with autonomous experience accumulation and sophisticated tool orchestration for spatial reasoning. First, \ModelNameL~structures experience into a two-level Hierarchical Memory Tree: the macro-level \textit{scene memory} abstracts environment-wide tool preferences, while the micro-level \textit{task memory} meticulously schedules specialist selection and execution sequencing for fine-grained spatial tasks. This architecture mitigates blind trial-and-error by enabling context-aware memory retrieval. Second, \ModelNameL~implements a dynamic dual-state memory mechanism: a \textit{construction state} that indexes successful trajectories alongside empirical performance statistics, and an \textit{evolution state} that diagnostically untangles failed executions. By analyzing reasoning traces and intermediate visual outputs, it proactively rectifies sub-optimal behaviors to synthesize superior specialist combinations.


Besides, existing spatial reasoning benchmarks suffer from quality issues (e.g., ambiguous references, truncated instances, and blurry objects) and provide limited coverage of scene and task types. To address this gap, we introduce \DatasetName, a benchmark for quantitative spatial reasoning covering object–object distance, object–camera distance, and object size across indoor and outdoor scenes, with rule-based filtering and manual verification to ensure data quality.

We conduct extensive experiments on multiple quantitative spatial reasoning benchmarks, including \DatasetName~, SpatialRGPT~\cite{cheng2024spatialrgpt}, and QSpatial-Bench~\cite{liao2024reasoning}, and qualitative spatial reasoning benchmarks CV-Bench~\cite{tong2024cambrian}. Our method consistently outperforms existing spatial specialist models as well as proprietary large-scale models, including GPT-5.4 and Gemini-3.1-Pro Preview. The hierarchical memory system improves performance over the memory-free variant, achieving an average gain of 9.1\%. These results demonstrate that our framework enhances agents’ ability to select and utilize tools in different scenarios.

In summary, our contributions are as follows:
\begin{itemize}[leftmargin=*]

\item We propose \ModelNameL, a hierarchical memory framework that integrates scene-level and task-level memory, while continuously optimizing suboptimal tool trajectories, enabling agents to select better external vision tools across diverse scenes and tasks.

\item We release \DatasetName, a well-curated benchmark for quantitative spatial reasoning, covering object-object distance estimation, object-camera distance estimation, and object size estimation in both indoor and outdoor environments.

\item Extensive experiments demonstrate that hierarchical memory design and the proposed trajectory optimization mechanism substantially improve agent performance across diverse spatial reasoning benchmarks.
\end{itemize}


%% file: section/related.tex
\vspace{-1ex}
\section{Related Work}
\vspace{-1ex}
\paragraph{VLMs Spatial Reasoning.}
Recent advances in VLMs have spurred increasing research into their spatial reasoning capabilities. Existing works typically frame spatial reasoning as a visual question answering task on image-text pairs. To address this task, current approaches follow two main technical routes. The first synthesizes training data from 3D datasets or simulated environments, as exemplified by SpatialRGPT~\cite{cheng2024spatialrgpt} and SpatialVLM~\cite{chen2024spatialvlm}, which generate templated spatial queries via data pipelines. The second incorporates geometric specialist modules into VLMs, such as VLM-3R~\cite{fan2025vlm} with CUT3R for 3D information injection and RoboRefer~\cite{zhou2026roborefer} with a depth encoder for depth feature extraction. However, both routes require either large-scale training data or architectural modifications, which limit interpretability and increase system complexity.

Rather than integrating spatial reasoning capabilities into the model itself, we enable VLMs to invoke external vision specialist models as tools for solving spatial reasoning tasks.
\paragraph{Tool-augmented Reasoning. }
Tool-augmented reasoning enables LLMs and VLMs to invoke external tools. In spatial reasoning, recent studies leverage vision specialist models to enhance understanding. One line of work trains VLMs to interact with external tools, such as SpaceTool~\cite{chen2025spacetools} and TIGeR~\cite{han2025tiger}, using supervised fine-tuning or reinforcement learning to improve tool-use capability. However, these methods demand large-scale training data and substantial computation. Another line comprises training-free approaches: VADAR~\cite{marsili2025visual} generates a Pythonic API with task-specific functions, while SpatialPIN~\cite{ma2024spatialpin} relies on a predefined tool pipeline for question answering. Yet, these training-free methods typically depend on a single vision specialist per task, with no exploration of selecting a suitable specialist from a candidate pool.

\paragraph{Memory-enhanced Reasoning.}
Memory has been widely adopted in agents to accumulate experience and improve generalization. Existing mechanisms fall into two paradigms: parametric memory, which stores knowledge in model parameters~\cite{chen2023fireact,yin2024agent}, and non-parametric memory, which leverages external databases or services for persistent context~\cite{packer2023memgpt,qian2025memorag}. However, their general framework is not tailored to spatial reasoning tasks, and their ability to select appropriate vision specialists under varying conditions remains underexplored. Our work falls within the non-parametric paradigm, extending memory-based agents to spatial reasoning by enabling adaptive specialist selection based on historical experience.

%% file: section/formulation.tex
\vspace{-1ex}
\section{Problem Formulation}
\vspace{-1ex}

We organize the heterogeneous tool space $\mathcal{T}$ into $P$ functional groups $\mathcal{G} = \{G_1, G_2, \ldots, G_P\}$. Each group $G_p$ aggregates multiple vision specialists that share identical functional objectives but exhibit complementary performance characteristics under varying scene dynamics. For example, the depth estimation group is formalized as $G_{\text{depth}} = \{\texttt{UniDepth}, \texttt{DepthPro}, \texttt{MoGe-2}\}$. The complete tool space is defined as the union of all functional clusters $\mathcal{T} = \bigcup_{p=1}^{P} G_p$, spanning vital vision capabilities such as segmentation, keypoint pointing, 2D/3D bounding box detection, and point cloud estimation (see Appendix~\ref{app:spatialmem} for full details).

We formulate quantitative spatial reasoning as a multi-turn, tool-augmented question answering task. Formally, given an image $\mathcal{I}$ and a spatial query $\mathcal{Q}$, the VLM agent iteratively reasons and orchestrates vision specialists from $\mathcal{T}$ across multiple execution turns to yield the final quantitative textual answer $\mathcal{A}$.

%% file: section/method.tex
\section{Method}
\begin{figure*}[t]
\begin{center}
\centerline{\includegraphics[width=\textwidth]{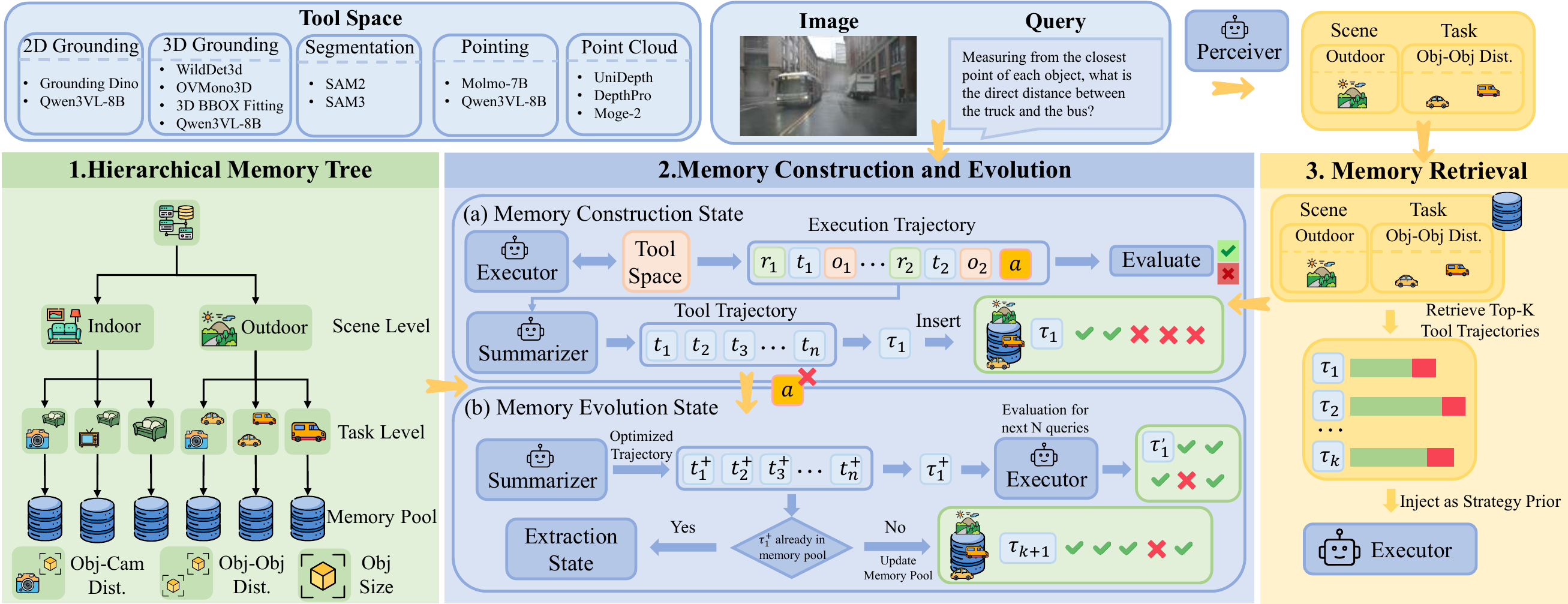}}
\caption{Overview of \ModelNameL, which consists of three components: (1) a hierarchical memory tree for scene-task memory organization, (2) a memory construction and evolution module for trajectory updating and optimization, and (3) a memory retrieval module for adaptive specialist selection in quantitative spatial reasoning.}
\vspace{-5ex}
\label{fig:main}
\end{center}
\end{figure*}
\subsection{SpatialMem}
Our framework is illustrated in Figure~\ref{fig:main}. \ModelNameL~is an orchestration framework designed to supercharge Vision-Language Model (VLM) agents with long-term memory for quantitative spatial reasoning. The framework consists of three components:
(1) A structured two-level memory tree that maps agent experiences into a joint environment-task space, ensuring structural isolation and preventing memory retrieval from cross-domain interference. (2) A continuous optimization loop consisting of a \textit{Construction State} to index empirical successes, and an \textit{Evolution State} that diagnoses failed trajectories to iteratively refine tool pipelines from execution feedback.
(3) An accuracy-driven ranking strategy that identifies elite tool execution chains and instantiates them as robust reasoning priors to guide adaptive tool orchestration.

\paragraph{Hierarchical Memory Tree}
To prevent semantic cross-contamination and ensure precision during experience invocation, \ModelNameL~ structures the agent’s accumulated experience into a two-level hierarchical memory tree.

At the first level, memories are partitioned by environmental dimension
$
s \in \mathcal{S} = \{\texttt{Indoor}, \texttt{Outdoor}\},
$
This macro-partitioning is physically grounded: indoor and outdoor environments differ significantly in depth ranges, lighting conditions, and geometric structures. These distinct characteristics directly influence the performance and selection of downstream vision specialists. At the second level, memories are further partitioned by task dimension
$
c \in \mathcal{C} = \{\texttt{OO Dist.}, \texttt{OC Dist.}, \texttt{Obj. Size}\},
$
which corresponds to object-object distance estimation, object-camera distance estimation, and object size estimation, respectively.

Each scene-task pair $(s,c)$ defines an independent memory pool $Mem_{(s,c)}$ that stores category-specific tool trajectories, enabling accurate retrieval while reducing cross-category interference.

\paragraph{Memory Construction and Evolution}

Given an image $I$ and a query $Q$,
the perception model
$\mathcal{M}_{\text{perceiver}}$ predicts the scene-task category pair $(s,c)$ for memory routing via a structured schema. Once routed, the memory framework operates under a dual-state dynamic process to continuously enrich and refine the corresponding memory pool $Mem_{(s,c)}$ .

\textbf{Construction State}
During the construction state, the execution agent
$\mathcal{M}_{\text{executor}}$
interacts with the tool space $\mathcal{T}$ to generate an execution trajectory
\[
\Gamma = \{r_1,t_1,o_1,\dots,r_n,t_n,o_n,a\},
\]
which contains reasoning traces, tool invocations, visual tool outputs, and the final answer.

Upon verifying the execution result against the ground truth, an empirical correctness score
$
r \in \{0,1\}
$ is assigned

The summarization model
$\mathcal{M}_{\text{summarizer}}$
then extracts a compact tool trajectory using a structured prompt from $\Gamma$.
\[
\tau = \{t_1,t_2,\dots,t_n\},
\]
which abstracts the pure sequence of specialist invocations. For a newly discovered trajectory $\tau$, its empirical entry is initialized in c as $\mathbf{r}_{\tau}^{(s,c)} = [r]$. For recurring trajectories, the historical performance vector is appended:
\[
\mathbf{r}_{\tau}^{(s,c)}
\leftarrow
\mathbf{r}_{\tau}^{(s,c)} \cup \{r\}.
\]

\textbf{Evolution State}
When execution fails $(r=0)$, \ModelNameL~triggers an evolution mechanism to optimize sub-optimal behaviors. Instead of discarding failed trajectories, $\mathcal{M}_{\text{summarizer}}$ diagnoses the execution failure by analyzing the complete trajectory $\Gamma$, including erroneous reasoning states and imperfect intermediate visual outputs. Based on this mistake analysis, the summarizer synthesizes an optimized candidate trajectory $\tau^{+}$ via category-consistent tool substitution and re-ordering. If $\tau^{+}$ does not already exist in $Mem_{(s,c)}$, the framework introduces $\tau^{+}$ into an evolution stage. During this phase, trajectory optimization is temporarily disabled, and $\tau^{+}$ is strictly leveraged as the primary retrieval prior for the next $N$ queries within the same $(s,c)$ category. After the evolution period, the framework records the empirical score sequence
\[
\mathbf{r}_{\tau^{+}}^{(s,c)}
=
[r_1,r_2,\dots,r_N]
\]
and merges the tool trajectory into the confirmed memory pool.

To avoid interference from multiple unverified strategies, each memory pool maintains at most one in-evolution trajectory at a time. The hyperparameter $N$ controls the tradeoff between estimation reliability and exploration speed: larger values provide more stable performance estimates before promotion into the confirmed pool, while smaller values enable faster discovery of improved tool pipelines. The prompt for the $\mathcal{M}_{\text{perceiver}}$ and $\mathcal{M}_{\text{summarizer}}$ is provided in the Appendix~\ref{app:prompt}.

\paragraph{Memory Retrieval}
When a new instance is routed to $Mem_{(s,c)}$, \ModelNameL~retrieves the top-$K$ elite tool trajectories from the confirmed memory pool, sorted by their empirical mean accuracy:
$\bar{r}_{\tau}^{(s,c)}.$
The retrieved tool chains are injected into the context of $\mathcal{M}_{\text{executor}}$ as strategy priors, enabling the agent to execute complex specialist pipelines tailored to the underlying spatial geometry

\subsection{Data Construction}

To address quality issues in existing spatial reasoning benchmarks, including blurry objects, ambiguous references, and limited diversity, we introduce \DatasetName, a curated quantitative benchmark consisting of 658 questions covering object size estimation, object-object distance estimation, and object-camera distance estimation. The dataset is built from Omni3D~\cite{brazil2023omni3d}, including SUNRGBD indoor scenes and KITTI/nuScenes outdoor scenes, spanning 81 object categories with physical scales from 20 cm to 50 m.

Following previous work~\cite{fan2025vlm, yang2025thinking}, we retain only object instances whose category appears exactly once in each image to avoid reference ambiguity. For object-object distance estimation, we also include same-category pairs (e.g., distance between two chairs). For visual quality, we apply a visibility threshold of 0.9 and discard truncated instances with bounding boxes intersecting image boundaries. We construct raw spatial QA pairs from the filtered Omni3D annotations. These raw data exhibit two systematic biases: category imbalance (frequent categories dominate) and value-range skew (short distances over‑represented). We mitigate these via stratified sampling over the ground‑truth value range grouped by category, ensuring balanced and diverse coverage. After automatic filtering, issues such as blurry objects and ambiguous references persist. We therefore instruct human annotators to manually refine the dataset, ensuring high annotation quality and spatial clarity.

\input{Tables/main_table}

%% file: Tables/main_table.tex
\begin{table*}[t]
\caption{Evaluation results of all methods on the \DatasetName. Performance is evaluated using the success rate, defined as the percentage of predictions falling within $[0.80\times, 1.25\times]$ of the ground-truth value.}
\vspace{-2pt}
\small
\resizebox{1\textwidth}{!}{
\begin{tabular}{lccccccccc}
\toprule
\multirow{2}{*}{\textbf{Algorithms}} 
& \multirow{2}{*}{\textbf{Setting}} 
& \multicolumn{4}{c}{\textbf{MetricBench-Indoor}} 
& \multicolumn{4}{c}{\textbf{MetricBench-Outdoor}} \\
\cmidrule(lr){3-6}
\cmidrule(lr){7-10}
&& \textbf{OO Dist.} & \textbf{OC Dist.}& \textbf{Obj. Size}& \textbf{Average} 
& \textbf{OO Dist.} & \textbf{OC Dist.}& \textbf{Obj. Size}& \textbf{Average} \\
\midrule
\rowcolor[HTML]{F5F5F5}
\multicolumn{10}{c}{\textit{Proprietary Models}} \\
GLM-4.6V &-&40.8&34.7&24.5&33.3&20.9&28.6&55.2 &34.9 \\    
GPT-5.4 &-&17.1&30.6&27.4&25.0&20.9&31.4&67.6 & 39.9 \\
Gemini-3.1-Pro-Preview &-&37.3&28.6&24.4&30.1&16.0&38.2&66.7 &40.3  \\
\midrule
\rowcolor[HTML]{F5F5F5}
\multicolumn{10}{c}{\textit{General Open-Source Models}} \\

Qwen3VL-8B &-&16.0&16.7&12.9&15.2&19.1&17.2&66.0&34.1\\
Qwen3VL-30B-A3B &-&28.0&17.6&26.0&23.8&13.6&33.1&54.6&33.7\\
InternVL3.5-8B &-&27.6&5.8&15.3&16.2&13.6&9.3&53.3&25.4\\

\rowcolor[HTML]{F5F5F5}
\multicolumn{10}{c}{\textit{Spatial VLMs}} \\ 
SpatialRGPT-8B &-&23.7&19.0&15.1&19.2&8.2&1.4&47.1&18.9\\

RoboRefer-8B-SFT&-&19.7&71.9&5.7&32.4&29.1&47.1&12.6&29.6 \\
SD-VLM &-&22.4&28.1&11.3&20.6&7.7&0&4.8&4.1 \\
\rowcolor[HTML]{F5F5F5}
\multicolumn{10}{c}{\textit{Agentic
}} \\ 
No Memory&-& 36.8&79.3&31.1&49.0&52.7&74.2&80.0&68.9\\ 
ReasoningBank&Online-VLM&39.4&75.2&29.2&47.9&51.8&72.8&75.2&66.6 \\
ReasoningBank&Online-Oracle&43.4&76.8&32.0&50.7&\second{53.6}&75.0&73.3&67.3\\
ReasoningBank&Offline&44.7&77.6&33.9&52.0&51.8&74.2&75.2&67.0 \\
ReMe&Online-VLM &40.7&76.8&30.1&49.6&50.0&69.2&71.4&63.5 \\
ReMe&Online-Oracle&40.7&80.9&33.0&51.5&50.9&72.8&78.0&64.8\\
blue{ReMe}&Offline&42.1&79.3&33.9&51.7&51.8&74.2&76.1&67.3 \\
\ModelNameL &Online-VLM&43.4&79.3&29.2&50.6&50.0&74.2&74.2&66.1\\
\ModelNameL &Online-Oracle&\second{53.9}&\second{85.9}&\second{37.7}&\second{59.1}&53.3&\second{77.1}&\second{81.9}&\second{70.7}\\
\ModelNameL&Offline&\best{57.8}&\best{89.2}&\best{38.6}&\best{61.8}&\best{62.7}&\best{79.2}&\best{84.7}&\best{75.5} \\
\bottomrule
\end{tabular}}
\vspace{-3ex}
\label{Table:overall_result_bench}
\end{table*}

%% file: section/experiment.tex
\section{Experiment}

\paragraph{Setup}
We evaluate \ModelNameL~on multiple quantitative spatial reasoning benchmarks, including Metric-Bench, SpatialRGPT~\cite{cheng2024spatialrgpt}, QSpatial-ScanNet~\cite{liao2024reasoning},  out of domain test setQSpatial++~\cite{liao2024reasoning} and multi-image benchmark MMSI-Bench~\cite{yang2025mmsi}, as well as the qualitative benchmark CV-Bench~\cite{tong2024cambrian}. Our agent executor, perceiver, and summarizer are built upon GPT-5.4-Nano. For each memory-based method, we evaluate two self-evolving settings: \textbf{online} and \textbf{offline}. In the \textbf{online} setting, the agent dynamically updates the memory tree using execution feedback. We consider \textbf{Online w/ Oracle Feedback}, which uses ground-truth verification, and \textbf{Online w/ VLM Feedback}, which relies on a VLM judge without ground-truth labels. In the offline setting, the memory system first accumulates experience from a training set, which is constructed from the Omni3D training split following the same pipeline as \DatasetName, and is then evaluated on \DatasetName. The details of all benchmarks and the training set are shown in Appendix~\ref{app:dataset}.


\paragraph{Baselines}
We compare our approach against four categories of baselines: (1) proprietary models, including GLM‑4.6V~\cite{vteam2025glm45vglm41vthinkingversatilemultimodal}, GPT‑5.4, and Gemini‑3.1‑Pro-Preview; (2) general‑purpose open‑source models, including Qwen‑3‑VL~\cite{bai2025qwen3} and InternVL3.5~\cite{wang2025internvl3}; and (3) specialist spatial reasoning models, including SpatialRGPT~\cite{cheng2024spatialrgpt}, RoboRefer~\cite{zhou2026roborefer}, SD‑VLM~\cite{chen2026sd}, and MM‑Spatial~\cite{daxberger2025mm}; (4) agentic reasoning methods, including variants without memory as well as memory-integrated approaches such as ReasoningBank~\cite{ouyang2025reasoningbank} and ReMe~\cite{cao2025remember}.

\paragraph{Evaluation} We follow SpatialRGPT~\cite{cheng2024spatialrgpt} and adopt a success criterion where predictions within $0.80\times$ to $1.25\times$ of the ground-truth value are considered correct across all quantitative benchmarks. For qualitative benchmarks, we use multiple-choice evaluation and report classification accuracy. The details of the implementations and evaluation are in Appendix~\ref{app:detail}.

\input{Tables/spatialrgpt}
\subsection{Quantitative Reasoning Evaluation}

\paragraph{Results on \DatasetName} 
The results on \DatasetName~are presented in Table~\ref{Table:overall_result_bench}. We have the following observations:
 
\begin{itemize}[leftmargin=*]
    \item \textbf{Agentic vs. Non-Agentic Methods.} 
    Agentic methods consistently outperform non-agentic baselines, achieving about 20-30\% higher success rates than pure VLMs, highlighting the importance of iterative tool use and reasoning for quantitative spatial understanding. \ModelNameL~achieves the best overall performance, surpassing both memory-free and existing memory-integrated methods by up to 9.7\%, validating the benefits of hierarchical memory and evolution design for adaptive tool selection. The offline variant further improves the result, outperforming the online setting by approximately 3\% on average.
 
     %
    \item \textbf{Indoor vs. Outdoor.} Baseline methods achieve on average over 30\% higher success rates on outdoor object size estimation than in indoor scenarios. This is mainly because outdoor scenes contain objects with relatively fixed metric priors, such as trucks, cars, and humans, allowing VLMs to leverage strong semantic inductive biases to infer approximate object sizes. In contrast, indoor objects exhibit much larger scale variations, making such semantic priors less reliable.
     
    \item  \textbf{Size Estimation vs. Distance Estimation.} Object size estimation significantly outperforms both distance estimation tasks in outdoor scenes. Notably, object-object distance estimation success rates for proprietary and open-source models remain below 20\% in outdoor scenes. This indicates that current VLMs rely on semantic priors for coarse scale estimation but struggle with precise distance measurement, which demands stronger geometric reasoning.
    
 
    \item \textbf{Online vs. Offline.} We observe that incorporating VLM feedback degrades the performance of all agentic memory methods, highlighting the limitations of current models in evaluating fine-grained spatial reasoning tasks. In contrast, with oracle feedback, SpatialMem outperforms the best competing memory method by 8.4\% and 3.4\%, respectively. With offline experience accumulation, SpatialMem surpasses the best baseline by 10.1\% and 8.2\%. These results demonstrate the effectiveness of its hierarchical memory and trajectory-level evolution mechanism.


\end{itemize}

\input{Tables/ablation_study}

\paragraph{Results on other Benchmarks}

The results on SpatialRGPT-Quantitative, QSpatial-ScanNet, QSpatial++, and MMSI-Bench are shown in Table~\ref{Table:overall_result_other}. \ModelNameL~consistently outperforms baselines across most quantitative reasoning tasks.

On SpatialRGPT-Quantitative, spatial VLMs perform well on size and horizontal distance estimation. For instance, MM-Spatial-3B achieves the best results on height estimation and horizontal distance estimation with scores of 83.4 and 82.0, respectively, suggesting that specialized training enhances fine-grained geometric perception.

However, their performance does not generalize consistently across benchmarks. SpatialRGPT-8B and RoboRefer-8B degrade substantially on QSpatial-ScanNet and QSpatial++, falling below 20\% success rates on object-object distance estimation. This indicates limited robustness under distribution shifts, especially on QSpatial++, a more challenging out-of-domain benchmark.

In contrast, \ModelNameL~ improves width estimation from 57.9\% to 84.2\% compared to the best existing baselines, vertical distance estimation from 63.2\% to 81.1\%, and QSpatial-ScanNet object-object distance estimation from 41.2\% to 69.6\%. Similar large-margin gains are observed on object size estimation and QSpatial++ and MMSI-Bench. These consistent improvements demonstrate that our agentic memory mechanism effectively transfers spatial experiences across environments and reasoning tasks.

\subsection{Ablation Study}

We evaluate the contribution of each core component in \ModelNameL~(offline)on \DatasetName. Table~\ref{Table:ablation} reports the results under four ablated configurations: (1) \textit{w/o Memory}: remove the entire memory framework; (2) \textit{w/o Domain-Specific Structure}; (3) \textit{w/o Scene-Level Memory}: eliminate scene-level memory; (4) \textit{w/o Task-Level Memory}: disable task-level memory; (5) \textit{w/o Trajectory Optimization}: deactivate trajectory refinement.

\paragraph{Impact of Hierarchical Memory}
Removing the domain-specific structure reduces success rates from 61.8\% to 51.7\% on indoor scenes and from 75.5\% to 65.5\% on outdoor scenes, demonstrating the importance of domain-aware memory organization. Scene-level memory provides moderate improvements, with removal decreasing performance to 58.2\% indoors and 72.3\% outdoors, as different environments require different specialists. Task-level memory is more critical, where its removal reduces performance to 51.6\% indoors and 65.5\% outdoors due to increased cross-task interference.

\begin{figure*}[h]
\begin{center}
\centerline{\includegraphics[width=\textwidth]{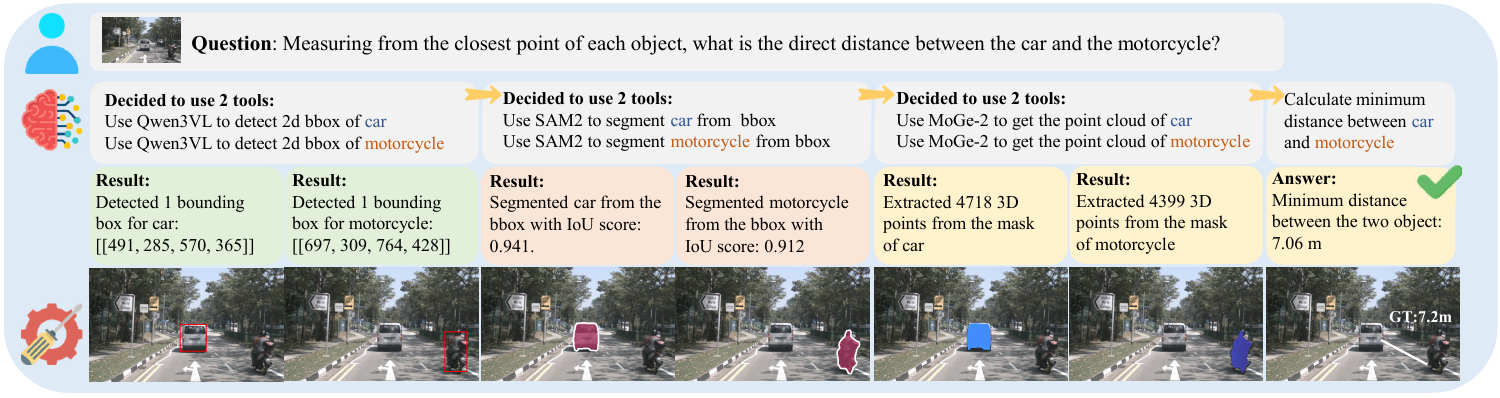}}
\caption{Example of \ModelNameL~solving a quantitative spatial reasoning task. The agent localizes objects, segments the detected boxes, extracts point clouds from the masks, and computes inter-object distances.}
\vspace{-4ex}
\label{fig:case_study}
\end{center}
\end{figure*}

\paragraph{Impact of Trajectory Optimization}
We further evaluate the memory evolution module by disabling trajectory optimization. In this setting, the framework stores historical trajectories without failure analysis or optimized trials. This causes a clear performance drop: average success rate falls to 49.0\% (indoor) and 67.9\% (outdoor). Notably, indoor object-object distance estimation decreases from 57.8\% to 33.3\%. These results confirm that merely accumulating trajectories is insufficient; optimization and trial-based validation are essential for continuous memory evolution and correcting suboptimal selections. Further results are shown in the Appendix~\ref{app:ablation}.

\input{Tables/spatial_reasoning}

\subsection{Qualitative Reasoning Evaluation}

We further evaluate qualitative spatial reasoning on CV-Bench (Table~\ref{Table:cv_bench}). Open-source VLMs already perform strongly on this benchmark. For example, Qwen3VL-8B achieves 97.2\% on both indoor depth and distance estimation, approaching specialized spatial VLMs, indicating that general-purpose VLMs have robust qualitative spatial understanding. Specialized models like MM-Spatial-3B and RoboRefer obtain the best results, yet their improvements over strong open-source baselines are marginal, within 1–3 points. Our method improves precise quantitative reasoning while maintaining competitive qualitative performance.

\begin{figure}[t]
\centerline{\includegraphics[width=\columnwidth]{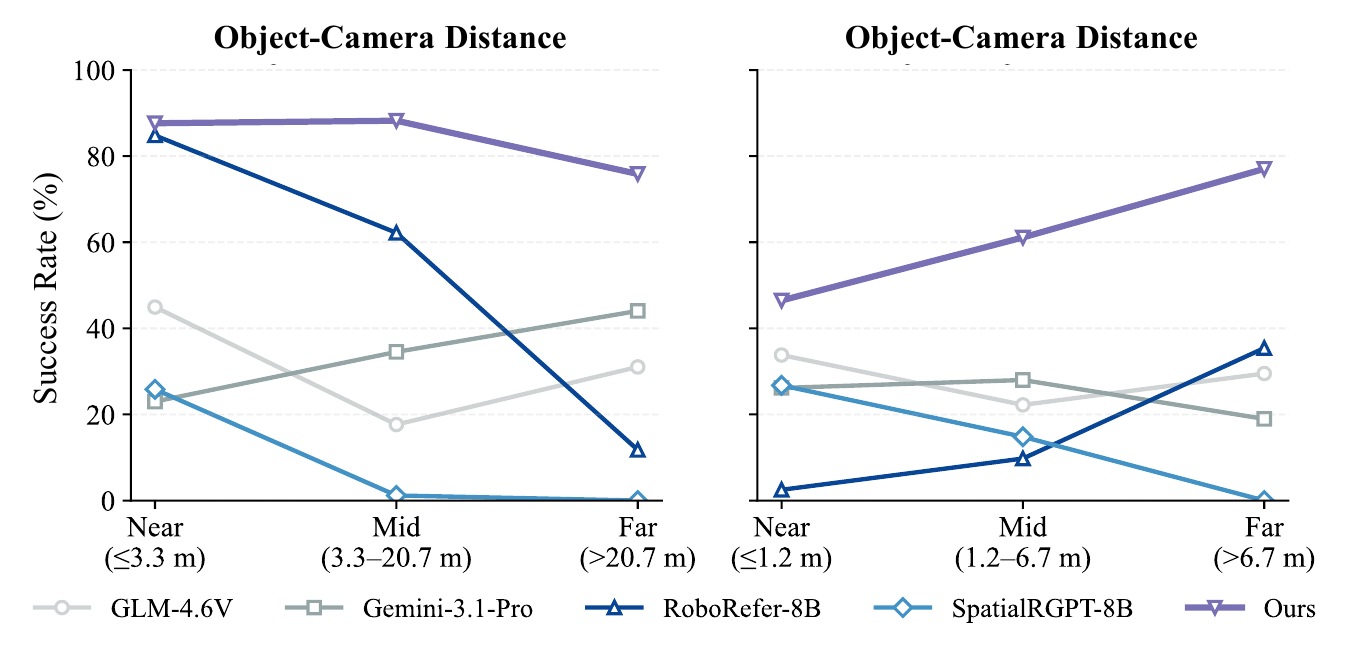}}
\caption{Method results across scaled distance. Success rates are shown for different models across "Near", "Mid", and "Far" distance categories.}
\label{fig:metirc_analysis}

\end{figure}

\subsection{Robustness Analysis}

We further analyze model robustness across physical distance ranges. For object-camera distance estimation, samples are evenly divided into Near, Mid, and Far groups using thresholds of $1.2\,\text{m}$ and $6.7\,\text{m}$. For object-object distance estimation, we uses thresholds of $3.3\,\text{m}$ and $20.7\,\text{m}$.

Figure~\ref{fig:metirc_analysis} shows that existing spatial reasoning models degrade significantly as physical scale increases. In object-camera estimation, RoboRefer-8B performs competitively in near-range but drops sharply from 84.5\% to 12.0\% in far-range, indicating limited robustness under large depth variations. SpatialRGPT-8B collapses to nearly 0\% beyond near-range. This degradation stems from training data dominated by near-range samples. In contrast, \ModelNameL~maintains strong performance across all scales, achieving over 75\% success rate even in far-range estimation.

A similar trend holds for object-object distance estimation. Most baselines struggle as object distance increases, especially in far-range scenarios requiring precise grounding and geometric reasoning. By contrast, \ModelNameL~exhibits steadily improving performance from near to far ranges, achieving the best robustness. These results demonstrate that memory-driven specialist selection enables robust reasoning across diverse geometric conditions and large-scale physical variations.

\subsection{Case Study}

Figure~\ref{fig:case_study} shows a reasoning trajectory of \ModelNameL~on a distance estimation task. Given an image with a car and a motorcycle, the agent first detects their 2D bounding boxes using Qwen3VL, then segments precise instance masks with SAM2, and finally extracts 3D point clouds using MoGe-2. The minimum Euclidean distance between the two point clouds is $7.06\,\text{m}$. This example illustrates how \ModelNameL~selects better vision specialists via memory for quantitative spatial reasoning.

%% file: Tables/spatialrgpt.tex
\begin{table*}[t]
\caption{Evaluation results of all methods on quantitative reasoning benchmark. Performance is evaluated using the success rate, defined as the percentage of predictions falling within $[0.80\times, 1.25\times]$ of the ground-truth value.}
\vspace{-2pt}
\small
\resizebox{1\textwidth}{!}{
\begin{tabular}{lcccccccccc}
\toprule
\multirow{2}{*}{\textbf{Algorithms}} 
& \multirow{2}{*}{\textbf{Setting}} 
&\multicolumn{5}{c}{\textbf{SpatialRGPT-Quantitative}} & \multicolumn{2}{c}{\textbf{QSpatial-ScanNet}}& \multicolumn{1}{c}{\textbf{QSpatial++}}& \multicolumn{1}{c}{\textbf{MMSI-Bench}}\\
\cmidrule(lr){3-7}
\cmidrule(lr){8-9}
\cmidrule(lr){10-10}
\cmidrule(lr){11-11}

&& \textbf{Width} & \textbf{Height}& \textbf{Dist.} 
& \textbf{Hor. Dist.} & \textbf{Ver. Dist.} & \textbf{OO Dist.} & \textbf{Obj. Size} & \textbf{OO Dis.}& \textbf{Meas.} \\
\midrule
\rowcolor[HTML]{F5F5F5}
\multicolumn{11}{c}{\textit{Proprietary Models}} \\
GLM-4.6V &-&43.6&58.6&23.0&23.0&18.9&32.8&60.0&34.7&37.5\\    
GPT-5.4   &-&57.9&66.9&10.1&14.8&7.5&36.8&60.0&27.7&51.6\\
Gemini-3.1-Pro-Preview  &-&55.0&61.7&18.2&18.9&5.9&41.2&\second{66.7}&38.8&40.6\\
\midrule
\rowcolor[HTML]{F5F5F5}
\multicolumn{11}{c}{\textit{General Open-Source Models}} \\

Qwen3VL-8B &-&27.0&40.6&31.1&31.9&18.8&22.1&45.2&15.8&39.1\\
Qwen3VL-30B-A3B &-&39.7&64.0&22.8&19.3&33.7&31.2&55.8&18.8 &35.9\\
InternVL3.5-8B &-&27.8&41.3&31.1&31.9&17.9&20.8&33.3&14.9&35.9 \\

\rowcolor[HTML]{F5F5F5}
\multicolumn{11}{c}{\textit{Spatial VLMs}} \\ 
SpatialRGPT-8B&-&48.9&61.7&45.9&\second{68.0}&56.6&17.6&42.2&22.8&29.6 \\
MM-Spatial-3B&-&55.6&\best{83.4}&59.5&\best{82.0}&63.2&-&-&-&- \\
RoboRefer-8B-SFT&-&17.3&60.1&49.3&37.7&28.3&16.4&35.9&10.9&28.1 \\
SD-VLM &-&26.2&42.2&25.2&37.0&35.8&13.0&29.5&19.2&29.6 \\
\rowcolor[HTML]{F5F5F5}
\multicolumn{11}{c}{\textit{Agentic
}} \\ 

No Memory&-&76.6&68.4&59.5&39.3&\second{77.3}&54.4&57.7&\second{49.5}&51.5\\

ReasoningBank&Online-Oracle&75.9&71.4&64.1&40.9&76.4&47.2&42.2&49.5&50.0\\
ReMe&Online-Oracle&78.1&73.6&62.8&41.8&75.4&56.0&55.5&47.5&46.8\\
\ModelNameL& Online-VLM & 73.6 & 66.1 &56.7&37.7 &73.5&48.0&44.4&44.5&46.8\\
\ModelNameL &Online-Oracle&\second{78.9}&79.7&\second{64.8}&46.7&\best{81.1}&\second{57.6}&51.0&\second{49.5}&\second{53.2}\\
\ModelNameL&Offline&\best{84.2}&\second{80.4}&\best{63.5}&44.2&\best{81.1}&\best{69.6}&\best{75.5}&\best{54.4}&\best{57.8}  \\
\bottomrule
\end{tabular}}
\vspace{-1ex}
\label{Table:overall_result_other}
\end{table*}

%% file: Tables/ablation_study.tex


\begin{table*}[t]
\caption{Ablation study on \ModelNameL(offline). Performance is evaluated using the success rate, defined as the percentage of predictions falling within $[0.80\times, 1.25\times]$ of the ground-truth value.}
\vspace{-2pt}
\small
\resizebox{1\textwidth}{!}{
\begin{tabular}{lcccccccc}
\toprule
\multirow{2}{*}{\textbf{Algorithms}} 
& \multicolumn{4}{c}{\textbf{\DatasetName-Indoor}} 
& \multicolumn{4}{c}{\textbf{\DatasetName-Outdoor}} \\
\cmidrule(lr){2-5}
\cmidrule(lr){6-9}
& \textbf{OO Dist.} & \textbf{OC Dist.}& \textbf{Object Size}& \textbf{Average} 
& \textbf{OO Dist.} & \textbf{OC Dist.}& \textbf{Object Size}& \textbf{Average} \\
\midrule

w/o Memory & 36.8&79.3&31.1&49.0&52.7&74.2&80.0&68.9\\  
w/o Domain-specific Structure&39.4&80.1&35.8&51.7&42.7&74.2&76.1&64.3 \\
w/o Scene-Level Memory &\second{55.2}&\second{81.8}&\second{37.7}&\second{58.2}&\second{57.2}&\second{77.8}&\second{81.9}&\second{72.3} \\
w/o Task-Level Memory &38.1&80.1&36.7&51.6&42.7&75.7&78.1&65.5 \\
w/o Trajectory Optimization&33.3&\second{81.8}&32.0&49.0&50.0&75.7&78.1&67.9 \\
\midrule

\ModelNameL&\best{57.8}&\best{89.2}&\best{38.6}&\best{61.8}&\best{62.7}&\best{79.2}&\best{84.7}&\best{75.5} \\
\bottomrule
\end{tabular}}
\vspace{-1ex}
\label{Table:ablation}
\end{table*}



%% file: Tables/spatial_reasoning.tex
\begin{table}[t]

\caption{Evaluation results of open-source models, spatial VLMs and our methods on the qualitative spatial reasoning benchmark CV-Bench.}
\vspace{-2pt}
\small
\resizebox{1\columnwidth}{!}{
\begin{tabular}{lcccc}
\toprule
\multirow{2}{*}{\textbf{Algorithms}} 
& \multicolumn{2}{c}{\textbf{CV-Bench-Depth}} 
& \multicolumn{2}{c}{\textbf{CV-Bench-Distance}} \\
\cmidrule(lr){2-3}
\cmidrule(lr){4-5}
 & \textbf{Indoor}& \textbf{Outdoor}& \textbf{Indoor} 
& \textbf{Outdoor}\\
\midrule

\rowcolor[HTML]{F5F5F5}
\multicolumn{5}{c}{\textit{General Open-Source Models}} \\

Qwen3VL-8B &97.2&\second{92.0}&97.2&76.8\\
Qwen3VL-30B-A3B & 98.0&90.0&95.7&72.2\\
InternVL3.5-8B &93.2&84.0&90.2&71.5\\

\rowcolor[HTML]{F5F5F5}
\multicolumn{5}{c}{\textit{Spatial VLMs}} \\ 
SpatialBot-3B&77.2&78.0&60.0&64.5\\
MM-Spatial-3B&\second{99.0}&\second{92.0}&\second{97.8}&\best{80.0} \\
RoboRefer-8B-SFT&\best{99.5}&\best{94.0}&\best{98.0}&\best{80.0}\\
\rowcolor[HTML]{F5F5F5}
\multicolumn{5}{c}{\textit{Agentic
}} \\ 
No Memory &95.1&88.4&92.0&\second{79.8} \\
\ModelNameL(offline)&97.6&88.4&94.5&\best{80.0}  \\

\bottomrule
\end{tabular}}
\vspace{-3ex}
\label{Table:cv_bench}
\end{table}

%% file: section/conclusion.tex
\vspace{-1ex}
\section{Conclusion}
\vspace{-1ex}
In this work, we proposed \ModelNameL, a hierarchical memory framework that enables agentic VLMs to accumulate experience for quantitative spatial reasoning. Its hierarchical memory tree selects better vision specialists for different scenarios and tasks, while a self-evolving strategy continuously refines the memory. We also introduced \DatasetName, a curated quantitative spatial reasoning benchmark. Experiments show that \ModelNameL~achieves state-of-the-art performance across multiple benchmarks and exhibits strong robustness across physical scales, demonstrating the effectiveness of memory-driven VLM agents for accurate quantitative reasoning.

%% file: section/appendix.tex
\section{Dataset Details}~\label{app:dataset}
\subsection{Dataset Details}

\begin{figure*}[t]
\begin{center}
\centerline{\includegraphics[width=\textwidth]{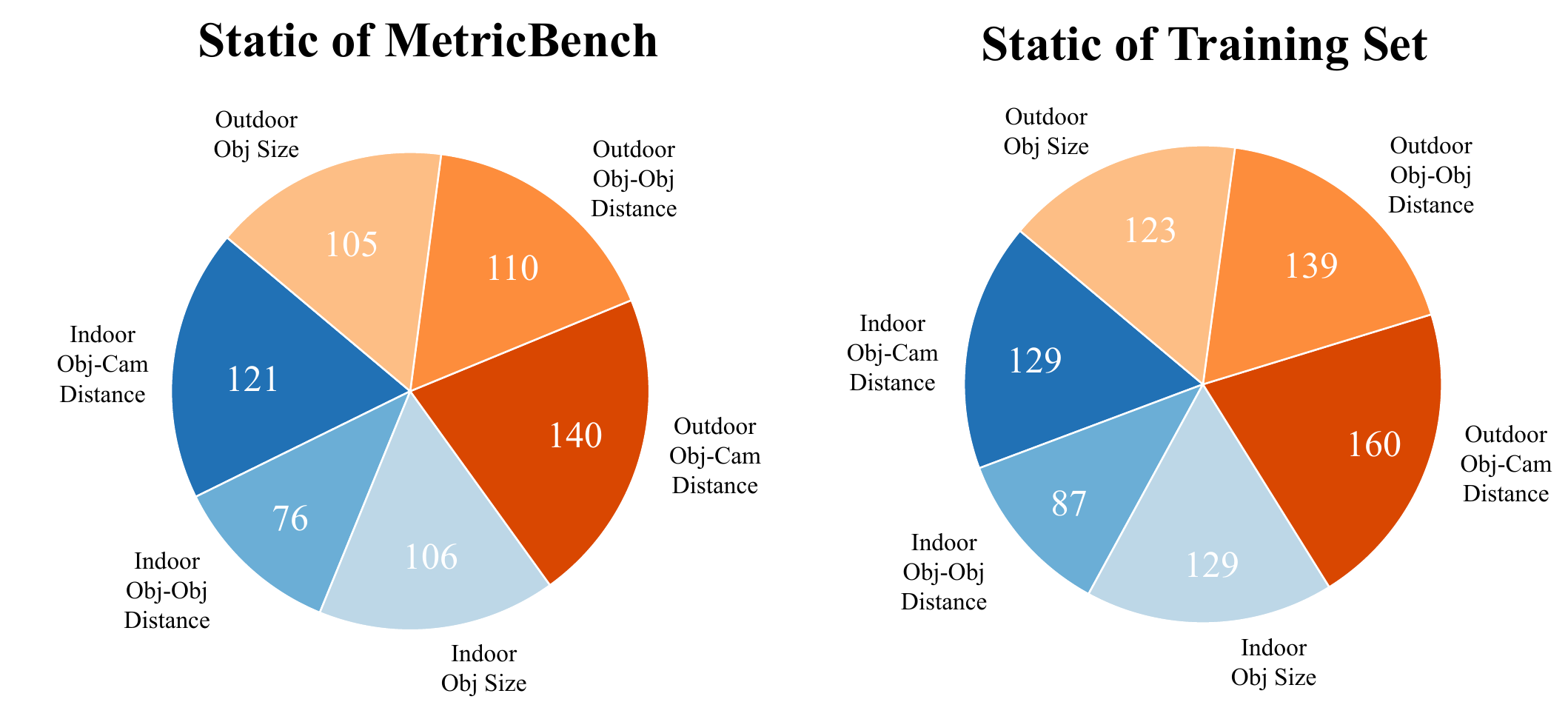}}
\caption{Composition of the \DatasetName (658 samples total). Samples are organized into three task types: object-camera distance, object-object distance, and object size under both indoor and outdoor environments. Indoor splits account for 303 samples and outdoor splits account for 355 samples.}
\label{fig:static}
\end{center}
\end{figure*}

\paragraph{\DatasetName}As illustrated in Figure~\ref{fig:static}, the \DatasetName~is divided into indoor (303 samples, 46.0\%) and outdoor (355 samples, 54.0\%) environments, providing a balanced coverage of both closed and open-scene settings. Each environment contains three task categories: object-to-camera distance estimation, object-to-object distance estimation, and object size estimation. This design ensures that the benchmark exercises a model's spatial understanding under varied depth cues, camera perspectives, and object interaction patterns, avoiding bias toward any single scene type or task.

\paragraph{Training Set} The statistics of the training set are shown on the right side of Figure~\ref{fig:static}. The training set is used to accumulate memory in the offline setting of our method and is further divided into indoor (345 samples) and outdoor (422 samples) subsets.

\paragraph{SpatialRGPT-Quantitative.}
We further evaluate our method on SpatialRGPT-Quantitative. SpatialRGPT-Quantitative includes additional metric reasoning tasks such as horizontal and vertical distance estimation. Moreover, it covers a substantially wider spatial scale range (0.15--41 m) compared with \DatasetName (0.3--25 m).

\paragraph{QSpatial++.}
We further evaluate our method on QSpatial++, an out-of-domain benchmark with respect to \DatasetName. Different from \DatasetName, whose images are derived from existing datasets, QSpatial++ images are manually collected from diverse real-world environments, and all spatial distances are physically measured rather than inferred from dataset annotations. In addition, QSpatial++ introduces a substantially different object and scene distribution, with 77 out of its 93 object categories being entirely disjoint from those in \DatasetName. These characteristics make QSpatial++ a more challenging testbed for evaluating cross-domain generalization.

\paragraph{MMSI-Bench.}
To further assess the robustness of our method under multi-image settings, we evaluate it on the measurement splits of the multi-image benchmark MMSI-Bench, which provides multi-object spatial configurations within a single scene.

\subsection{Dataset Comparison}
\input{Tables/dataset_comparison}
Table~\ref{Table:dataset_comparison} compares existing quantitative spatial reasoning benchmarks across scene diversity and task coverage. Existing benchmarks mainly focus on object-object distance estimation and object size estimation, while object-camera distance reasoning remains largely underexplored. In addition, several datasets only cover indoor scenes, limiting evaluation under diverse geometric conditions. In contrast, \DatasetName~provides balanced coverage across indoor and outdoor environments while simultaneously supporting object-object distance, object-camera distance, and object size estimation, enabling a more comprehensive evaluation of quantitative spatial reasoning capabilities.

\subsection{Dataset Construction}
\paragraph{Object-Object Distance Estimation}
Generates questions asking for the approximate Euclidean distance (in meters, numerical answer) between two target object instances in the image. \textbf{Generation Logic:} The 3D Euclidean distance is computed between the closest points on the two objects’ 3D bounding boxes in the camera coordinate system, where the 3D bounding boxes are derived from Omni3D.
\paragraph{Object-Camera Distance Estimation}
Generates questions asking for the approximate Euclidean distance (in meters, numerical answer) between the camera position and the closest point on the 3D bounding box of a target object instance in the image. \textbf{Generation Logic:} The 3D Euclidean distance is computed between the camera coordinates and the nearest point on the object’s 3D bounding box, which is derived from Omni3D.
\paragraph{Object-Size Estimation}
Generates questions asking for the approximate size (in meters, numerical answer) of a target object instance in the image. \textbf{Generation Logic:} The object size is directly obtained from the 3D bounding box annotations provided by Omni3D in the camera coordinate system.

\section{Details for Experiment Settings}~\label{app:detail}

\subsection{Implementation Details}

Our experiments are based on 8 NVIDIA A800 GPUs. In offline mode, the trial time $N$ is set to 5; in online mode, $N$ is set to 4 to balance reliability and exploration speed. The $K$ is set to 1 to select the most suitable experience. We use GPT-5.4 as the judge model for online w/ VLM Feedback. We adopt the Toolshed framework~\cite{chen2025spacetools}, which deploys multiple compute‑heavy tools alongside inference workloads and mitigates bottlenecks via resource and environment isolation for each tool instance. Further details are shown in Table~\ref{tab:implemen_details}.
\input{Tables/impletation_detail}

\subsection{Evaluation}
We leverage GPT-4o to assess the responses generated by different models in benchmark. The evaluation process involves providing a question, the correct answer, and the model’s response to the LLM. For quantitative questions, GPT-4o extracts numerical values from both the correct answer and the model’s response, converting them to the same unit (such as meters). For an estimated distance $\hat{d}$ and its ground truth value $d^{*}$, this ratio is calculated as
\[
\delta = \max \left( \frac{\hat{d}}{d^{*}}, \frac{d^{*}}{\hat{d}} \right).
\]

For the object size task in \DatasetName, we compute the mean scale ratio across the width, height, and length dimensions, and determine the success rate based on this aggregated ratio. Specifically, given the predicted object size
\[
\hat{\mathbf{s}} = (\hat{w}, \hat{h}, \hat{l})
\]
and the ground-truth size
\[
\mathbf{s}^{*} = (w^{*}, h^{*}, l^{*}),
\]
the scale ratio is defined as
\[
\begin{aligned}
\delta_{\text{size}}
=
\frac{1}{3} \Bigg(
&\max\left(\frac{\hat{w}}{w^{*}}, \frac{w^{*}}{\hat{w}}\right)
+
\max\left(\frac{\hat{h}}{h^{*}}, \frac{h^{*}}{\hat{h}}\right)
\\
&+
\max\left(\frac{\hat{l}}{l^{*}}, \frac{l^{*}}{\hat{l}}\right)
\Bigg).
\end{aligned}
\]
The success rate is then computed by applying a predefined threshold to $\delta_{\text{size}}$.
The prompts used for scoring by GPT-4 and other details are included Appendix~\ref{app:prompt}. We employ a threshold of 1.25 across all quantitative benchmark.

For qualitative questions, we use GPT-4 to determine if the model’s response aligns with the correct answer by assigning a score of 0 or 1. We provide prompts in.
\input{Tables/ablation_study_online}

\section{Additional Experiment}

\subsection{Comparison with Routing Baselines.}\label{app:routing}
\input{Tables/routing_baseline}
To investigate whether the performance gains of SpatialMem come from effective routing rather than merely accessing a large tool pool, we compare it with several routing baselines, including Random Trajectory, Majority Tool Pipeline, and Per-task Oracle Pipeline. As shown in Table~\ref{tab:routing_baselines}, SpatialMem consistently outperforms both non-adaptive routing strategies. Compared with Random Trajectory and Majority Tool Pipeline, SpatialMem achieves improvements of 14.8\% and 7.5\% on Indoor tasks, and 9.6\% and 7.6\% on Outdoor tasks, respectively. These results demonstrate that simply selecting arbitrary tools or relying on globally frequent tool choices is insufficient for quantitative spatial reasoning, where the optimal tool combination depends on both scene and task characteristics.

Per-task Oracle Pipeline achieves the best performance by exhaustively searching all candidate tool trajectories using task-specific evaluation, serving as an upper bound for routing performance. Nevertheless, SpatialMem approaches this oracle performance without requiring task-level exhaustive search, achieving 61.8\% and 75.5\% average success rates on Indoor and Outdoor settings, respectively. This indicates that the proposed hierarchical memory and trajectory evolution mechanism can effectively discover and reuse high-quality tool combinations through experience accumulation and retrieval.
\subsection{Ablation Study}\label{app:ablation}

\input{Tables/ablation_study_app}

We further conduct an ablation study on the domain‑specific structure and the self‑evolution mechanism. The results are shown in Table~\ref{Table:ablation_app}. After introducing the domain-specific memory structure, all methods benefit from improved experience organization, but our self-evolving mechanism achieves the largest performance gain, indicating that it can exploit structured memories more effectively than existing general-purpose agent-memory methods.

Table~\ref{Table:app_ablation} presents the ablation study of \ModelNameL~(online), demonstrating that both hierarchical memory design and trajectory optimization contribute substantially to quantitative spatial reasoning performance. Removing scene-level memory causes a notable performance drop on indoor benchmarks, indicating that environment-specific experience is critical for selecting suitable specialists under different geometric conditions. Removing task-level memory also degrades overall performance, showing the importance of modeling task-dependent tool selection and execution patterns. In addition, removing trajectory optimization leads to the largest performance decline, particularly on outdoor benchmarks, suggesting that iterative refinement of tool-use trajectories is essential for robust reasoning in complex large-scale scenes. Overall, the complete framework achieves the best average performance across most settings, validating the effectiveness of combining hierarchical memory with trajectory optimization.

\section{Hyperparameter Analysis}

\begin{figure*}[t]
\begin{center}
\centerline{\includegraphics[width=\textwidth]{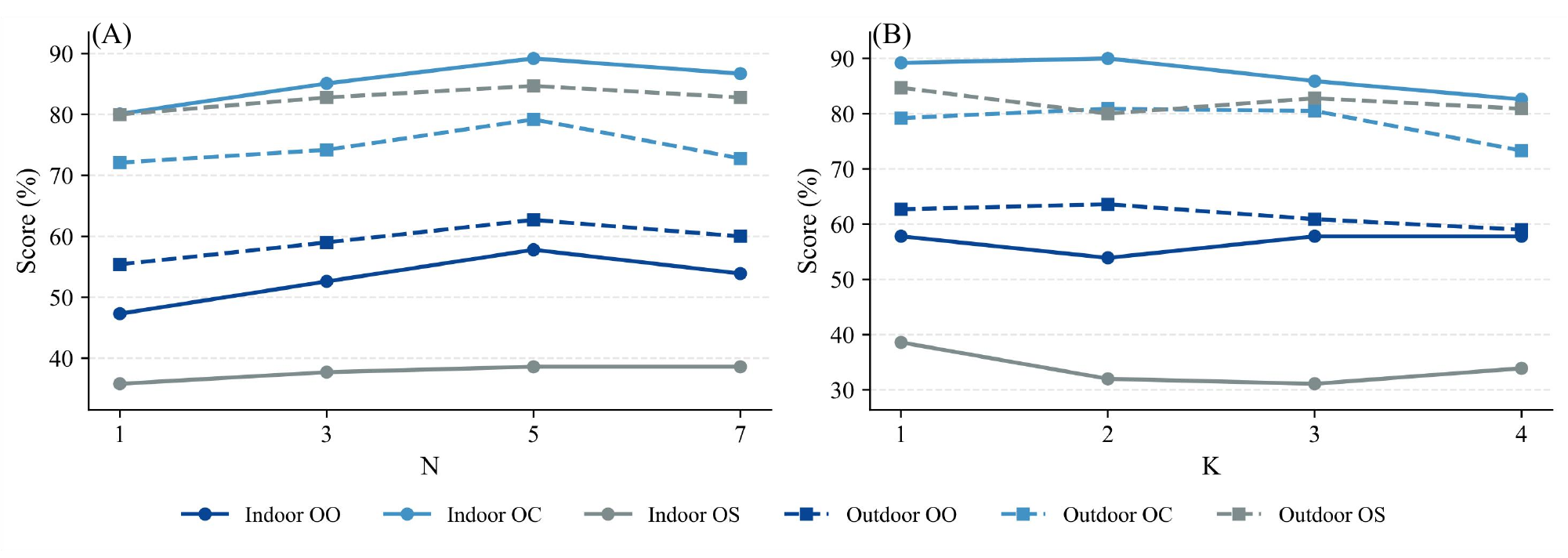}}
\caption{Effects of evolution period $N$ and candidate number $K$ on accuracy.}
\label{fig:hyper}
\end{center}
\end{figure*}

We further analyze the effects of two important hyperparameters, $N$ and $K$. As shown in Fig.~\ref{fig:hyper}(A), increasing the evolution period from $N=1$ to $N=5$ consistently improves performance, with the average accuracy increasing from 54.5\%/69.1\% to 61.8\%/75.5\% for indoor/outdoor settings, respectively. This improvement indicates that a moderate validation period allows the agent to accumulate sufficient evidence for evaluating evolved trajectories, thereby reducing the risk of promoting suboptimal strategies caused by noisy feedback. However, further increasing $N$ to 7 results in degraded performance. We attribute this decline to the reduced frequency of memory evolution, where overly long validation intervals limit opportunities for discovering and adopting better trajectory refinements.

Fig.~\ref{fig:hyper}(B) presents the impact of the retrieval size $K$. Retrieving only the top-1 trajectory achieves the best overall performance. Although a larger $K$ provides access to more historical experiences, it also introduces multiple tool chains with potentially different quality and applicability. These heterogeneous retrieval results may provide conflicting planning priors and increase the difficulty of selecting an optimal execution strategy. Therefore, retrieving the most effective trajectory offers a more reliable guidance signal for tool selection and pipeline planning.

\section{Details of \ModelNameL}\label{app:spatialmem}

\subsection{Provided Tools}
As shown in Table~\ref{tab:tool_pool}, we construct a diverse external vision tool pool covering multiple functional groups, including point cloud estimation, segmentation, 2D grounding, 3D grounding, and pointing. For point cloud estimation, we employ depth and geometry specialists such as \texttt{UniDepth}~\cite{piccinelli2024unidepth}, \texttt{DepthPro}~\cite{bochkovskiy2025depth}, and \texttt{MoGe-2}~\cite{wang2026moge}. For segmentation, we integrate \texttt{SAM2}~\cite{ravi2025sam} and \texttt{SAM3}~\cite{carion2025sam}. The 2D grounding module includes \texttt{GroundingDINO}~\cite{liu2024grounding} and \texttt{Qwen3VL-8B}~\cite{bai2025qwen3}, while the 3D grounding group combines \texttt{Qwen3VL-8B}~\cite{bai2025qwen3}, \texttt{WildDet3D}~\cite{huang2026wilddet3d}, \texttt{OVMono3D}~\cite{yao2024open}, and 3D box fitting methods. For pointing tasks, we adopt \texttt{Molmo-7B}~\cite{molmo2024}, \texttt{Qwen3VL-8B}~\cite{bai2025qwen3}, and 3D bounding-box fitting. These specialists provide complementary strengths under different scene conditions, enabling \ModelNameL~to dynamically select suitable tools for robust quantitative spatial reasoning.
\input{Tables/toollist}
\subsection{Detailed Tool APIs}
\input{section/detail_tool_api}

\input{section/case_study}

\subsection{Detailed trajectory analysis}

We further observe three patterns from the learned trajectories. First, DepthPro consistently performs best for outdoor object-camera estimation, whereas MoGe is preferred for object-object distance estimation, revealing clear scene-dependent tool preferences. Second, in the outdoor object-object distance estimation task, trajectories initialized by QwenVL achieve success rates of 0.48–0.64, compared with 0.12–0.27 for those initialized by GroundingDINO, demonstrating that the memory learns the reliability of alternative tool choices from historical experience. Third, 62.9\% of stored trajectories are improved through single-tool substitutions rather than complete pipeline redesign. These findings show that SpatialMem stores fine-grained tool usage experience rather than simply routing by scene or task category.
\FloatBarrier
\input{section/prompt}

%% file: Tables/dataset_comparison.tex
\begin{table*}[t]
\caption{Comparison of quantitative spatial reasoning benchmark.}
\centering
\vspace{-2pt}
\small
\resizebox{1\textwidth}{!}{
\begin{tabular}{lcccccc}
\toprule
\multirow{1}{*}{\textbf{Dataset}} & \textbf{Quantitative spatial questions}
& \textbf{Indoor}
&  \textbf{Outdoor}&  \textbf{Object-Object Distance} &\textbf{Object-Camera Distance} &\textbf{Object Size}\\
\midrule
\textbf{Spatial VQA}~\cite{chen2024spatialvlm} & 215&\checkmark&\checkmark&\checkmark&&\checkmark\\ 
\textbf{SpatialRGPT-Bench}~\cite{cheng2024spatialrgpt} &749&\checkmark&\checkmark&\checkmark&&\checkmark\\ 
\textbf{MSMU-Bench}~\cite{chen2026sd}& 299&\checkmark&&\checkmark&&\checkmark\\
\textbf{Q-Spatial}~\cite{liao2024reasoning} &271&\checkmark&\checkmark&\checkmark&&\checkmark\\    

\DatasetName&658&\checkmark&\checkmark&\checkmark&\checkmark&\checkmark \\

\bottomrule
\end{tabular}}
\vspace{-1ex}
\label{Table:dataset_comparison}
\end{table*}

%% file: Tables/impletation_detail.tex
\begin{table*}[h]
\centering
\caption{Detailed statistics of SpatialMem, including inference efficiency, tool usage, memory statistics, and evolution cost.}
\label{tab:implemen_details}
\resizebox{\linewidth}{!}{
\begin{tabular}{lcccccc}
\toprule
\textbf{Setting} & 
\textbf{Latency (s)} & 
\textbf{Avg. Tool Calls} & 
\textbf{Failure Rate} & 
\textbf{Memory Size} & 
\textbf{ Evolved Traj.} & 
\textbf{Cost (h)} \\
\midrule
Indoor OC  & 26.9 & 4.70 & 5.98\% & 18 & 10 & 0.8 \\
Indoor OO  & 32.9 & 8.20 & 1.28\% & 16 & 13 & 0.7 \\
Indoor OS  & 25.6 & 5.26 & 0.18\% & 31 & 18 & 0.7 \\
Outdoor OC & 27.8 & 5.62 & 1.14\% & 23 & 13 & 1.1 \\
Outdoor OO & 37.1 & 8.62 & 2.63\% & 21 & 14 & 0.9 \\
Outdoor OS & 4.8  & 1.00 & 0.94\% & 7  & 5  & 0.2 \\
\bottomrule
\end{tabular}
}
\end{table*}

%% file: Tables/ablation_study_online.tex
\begin{table*}[t]
\caption{Ablation study on \ModelNameL(Online-Oracle). Performance is evaluated using the success rate, defined as the percentage of predictions falling within $[0.80\times, 1.25\times]$ of the ground-truth value.}
\vspace{-2pt}
\small
\resizebox{1\textwidth}{!}{
\begin{tabular}{lcccccccc}
\toprule
\multirow{2}{*}{\textbf{Algorithms}} 
& \multicolumn{4}{c}{\textbf{Bench-Indoor}} 
& \multicolumn{4}{c}{\textbf{Bench-Outdoor}} \\
\cmidrule(lr){2-5}
\cmidrule(lr){6-9}
& \textbf{OO Dist.} & \textbf{OC Dist.}& \textbf{Object Size}& \textbf{Average} 
& \textbf{OO Dist.} & \textbf{OC Dist.}& \textbf{Object Size}& \textbf{Average} \\
\midrule
w/o Memory & 36.8&79.3&31.1&49.0&52.7&74.2&80.0&\second{68.9}\\   
w/o Scene-Level Memory &\second{46.0}&\second{80.1}&20.0&48.7&52.7&72.8&69.5&65.0 \\
w/o Task-Level Memory &\second{46.0} &79.3 &33.0 &52.7 &\best{56.3} &\best{78.5} &67.6 &67.4  \\
w/o Trajectory Optimization&43.4 &77.6 &\best{37.7} &\second{52.9} &\best{58.1} &76.4 &\second{79.0} &66.4  \\
\midrule

\ModelNameL(online)&\best{53.9}&\best{85.9}&\best{37.7}&\best{59.1}&53.3&\second{77.1}&\second{81.9}&\best{70.7} \\
\bottomrule
\end{tabular}}
\vspace{-1ex}
\label{Table:app_ablation}
\end{table*}

%% file: Tables/routing_baseline.tex
\begin{table*}[h]
\centering
\caption{Comparison with strong routing baselines.}
\label{tab:routing_baselines}
\resizebox{\linewidth}{!}{
\begin{tabular}{l|cccc|cccc}
\toprule
\multirow{2}{*}{Model} 
& \multicolumn{4}{c|}{Indoor} 
& \multicolumn{4}{c}{Outdoor} \\
\cmidrule(lr){2-5} \cmidrule(lr){6-9}
& OO & OC & OS & Avg. 
& OO & OC & OS & Avg. \\
\midrule
Random Trajectory 
& 34.2 & 77.6 & 29.2 & 47.0 
& 49.0 & 70.7 & 78.1 & 65.9 \\
Majority Tool Pipeline 
& 47.3 & 82.6 & 33.0 & 54.3 
& 51.8 & 75.0 & 77.1 & 67.9 \\
Per-task Oracle Pipeline 
& \textbf{60.5} & \textbf{91.7} & \textbf{49.0} & \textbf{67.0} 
& \textbf{63.6} & \textbf{82.1} & \textbf{89.5} & \textbf{78.4} \\
SpatialMem (Offline) 
& 57.8 & 89.2 & 38.6 & 61.8 
& 62.7 & 79.2 & 84.7 & 75.5 \\
\bottomrule
\end{tabular}
}
\end{table*}

%% file: Tables/ablation_study_app.tex


\begin{table*}[t]
\caption{Ablation study on \ModelNameL(offline). Performance is evaluated using the success rate, defined as the percentage of predictions falling within $[0.80\times, 1.25\times]$ of the ground-truth value.}
\vspace{-2pt}
\small
\resizebox{1\textwidth}{!}{
\begin{tabular}{lcccccccc}
\toprule
\multirow{2}{*}{\textbf{Algorithms}} 
& \multicolumn{4}{c}{\textbf{\DatasetName-Indoor}} 
& \multicolumn{4}{c}{\textbf{\DatasetName-Outdoor}} \\
\cmidrule(lr){2-5}
\cmidrule(lr){6-9}
& \textbf{OO Dist.} & \textbf{OC Dist.}& \textbf{Object Size}& \textbf{Average} 
& \textbf{OO Dist.} & \textbf{OC Dist.}& \textbf{Object Size}& \textbf{Average} \\
\midrule

w/o Memory & 36.8&79.3&31.1&49.0&52.7&74.2&80.0&68.9\\  
\textcolor{blue}{w/o Domain-specific Structure} &39.4&80.1&35.8&51.7&42.7&75.7&78.1&65.5 \\
w Domain-specific Structure + ReMe &\second{44.7}&\second{80.9}&\second{34.9}&\second{53.5}&\second{55.4}&\second{74.2}&\second{79.1}&\second{69.5} \\
w Domain-specific Structure + ReasoningBank &46.0&77.6&35.8&53.1&56.3&75.7&77.1&69.7 \\
\midrule

\ModelNameL&\best{57.8}&\best{89.2}&\best{38.6}&\best{61.8}&\best{62.7}&\best{79.2}&\best{84.7}&\best{75.5} \\
\bottomrule
\end{tabular}}
\vspace{-1ex}
\label{Table:ablation_app}
\end{table*}



%% file: Tables/toollist.tex
\begin{table*}[t]
\caption{Overview of the external vision tool pool used in \ModelNameL. 
Tools are organized into functional groups, where multiple specialists 
share similar functionality but exhibit complementary strengths under 
different scene conditions.}
\centering
\small
\begin{adjustbox}{width=0.95\textwidth}

\begin{tabular}{ll}
\toprule
\textbf{Functional Group} & \textbf{Vision Specialists} \\
\midrule

Point Cloud
& \texttt{UniDepth}~\cite{piccinelli2024unidepth},
 \texttt{DepthPro}~\cite{bochkovskiy2025depth},
\texttt{MoGe-2}~\cite{wang2026moge} \\
\midrule
Segmentation
&\texttt{SAM2}~\cite{ravi2025sam},
\texttt{SAM3}~\cite{carion2025sam} \\
\midrule
2D Grounding
&\texttt{GroundingDINO}~\cite{liu2024grounding},
\texttt{Qwen3VL-8B}~\cite{bai2025qwen3} \\
\midrule
3D Grounding
&\texttt{Qwen3VL-8B}~\cite{bai2025qwen3},
\texttt{WildDet3d}~\cite{huang2026wilddet3d},
\texttt{OVMono3d}~\cite{yao2024open},
\texttt{3D box fitting},  \\
\midrule
Pointing
&\texttt{Molmo-7B}~\cite{molmo2024},
\texttt{Qwen3-VL-8B}~\cite{bai2025qwen3},
3D BBox Fitting \\

\bottomrule
\end{tabular}
\end{adjustbox}
\label{tab:tool_pool}
\end{table*}

%% file: section/detail_tool_api.tex
Complete API of the computer vision includes:

\FloatBarrier
\begin{description}[style=nextline]
  \item[\texttt{sam2.segment\_from\_point(image, x, y, varname)}]
  Segment the object at a single pixel coordinate.
  \begin{itemize}
    \item \textbf{Inputs}:
      \begin{itemize}
        \item \texttt{image}: PIL.Image
        \item \texttt{x}: Normalized x-coordinate, float in \([0,1]\)
        \item \texttt{y}: Normalized y-coordinate, float in \([0,1]\)
        \item \texttt{varname}: Custom variable name to store the mask.
      \end{itemize}
    \item \textbf{Outputs}:
      \begin{itemize}
        \item \textbf{Raw value}: \texttt{dict} with \texttt{mask} (boolean \(H{\times}W\) \texttt{numpy.ndarray}) and \texttt{iou\_score} (float)
        \item \textbf{Text}: Reports the click location and IoU score
        \item \textbf{Image}: Original image with mask
        \item \textbf{Variables}: \texttt{segmentation\_mask} (boolean \(H{\times}W\) \texttt{numpy.ndarray})
      \end{itemize}
  \end{itemize}

  \item[\texttt{sam2.segment\_from\_bbox(image, bbox, varname)}]
  Segment the object at a single pixel coordinate.
  \begin{itemize}
    \item \textbf{Inputs}:
      \begin{itemize}
        \item \texttt{image}: PIL.Image
        \item \texttt{bbox}: Bounding box as list [x1, y1, x2, y2] in pixel coordinates
        \item \texttt{varname}: Custom variable name to store the mask.
      \end{itemize}
    \item \textbf{Outputs}:
      \begin{itemize}
        \item \textbf{Raw value}: \texttt{dict} with \texttt{mask} (boolean \(H{\times}W\) \texttt{numpy.ndarray}) and \texttt{iou\_score} (float)
        \item \textbf{Text}: Reports the click location and IoU score
        \item \textbf{Image}: Original image with mask
        \item \textbf{Variables}: \texttt{segmentation\_mask} (boolean \(H{\times}W\) \texttt{numpy.ndarray})
      \end{itemize}
  \end{itemize}
 \item[\texttt{sam3.segment\_from\_point(image, x, y, varname)}]
  Segment the object at a single pixel coordinate.
  \begin{itemize}
    \item \textbf{Inputs}:
      \begin{itemize}
        \item \texttt{image}: PIL.Image
        \item \texttt{x}: Normalized x-coordinate, float in \([0,1]\)
        \item \texttt{y}: Normalized y-coordinate, float in \([0,1]\)
        \item \texttt{varname}: Custom variable name to store the mask.
      \end{itemize}
    \item \textbf{Outputs}:
      \begin{itemize}
        \item \textbf{Raw value}: \texttt{dict} with \texttt{mask} (boolean \(H{\times}W\) \texttt{numpy.ndarray}) and \texttt{iou\_score} (float)
        \item \textbf{Text}: Reports the click location and IoU score
        \item \textbf{Image}: Original image with mask
        \item \textbf{Variables}: \texttt{segmentation\_mask} (boolean \(H{\times}W\) \texttt{numpy.ndarray})
      \end{itemize}
  \end{itemize}

  \item[\texttt{sam3.segment\_from\_bbox(image, bbox, varname)}]
  Segment the object at a single pixel coordinate.
  \begin{itemize}
    \item \textbf{Inputs}:
      \begin{itemize}
        \item \texttt{image}: PIL.Image
        \item \texttt{bbox}: Bounding box as list [x1, y1, x2, y2] in pixel coordinates
        \item \texttt{varname}: Custom variable name to store the mask.
      \end{itemize}
    \item \textbf{Outputs}:
      \begin{itemize}
        \item \textbf{Raw value}: \texttt{dict} with \texttt{mask} (boolean \(H{\times}W\) \texttt{numpy.ndarray}) and \texttt{iou\_score} (float)
        \item \textbf{Text}: Reports the click location and IoU score
        \item \textbf{Image}: Original image with mask
        \item \textbf{Variables}: \texttt{segmentation\_mask} (boolean \(H{\times}W\) \texttt{numpy.ndarray})
      \end{itemize}
  \end{itemize}

  \item[\texttt{molmo.detect\_points(image, obj\_name,num\_points)}]
  Identify instances of the named object by pointing to them with molmo.
  \begin{itemize}
    \item \textbf{Inputs}:
      \begin{itemize}
        \item \texttt{image}: PIL.Image
        \item \texttt{obj\_name}: string
        \item \texttt{num\_points}: Agent-selected maximum number of points
      \end{itemize}
    \item \textbf{Outputs}:
      \begin{itemize}
        \item \textbf{Raw value}: String serialization of a list of normalized point coordinates \((x,y) \in [0,1]^2\) 
        \item \textbf{Text}: Reports the object name, count, and the list of normalized points
        \item \textbf{Image}: Original image with red circular point markers at detected locations (white outlines)
        \item \textbf{Variables}: \texttt{\textit{<obj\_name>}\_detections} Stored variables: molmo\_obj\_name\_detections\_1, ...\_2, … (one per kept point; (x, y) in [0, 1]).
      \end{itemize}
  \end{itemize}

  \item[\texttt{Qwen3VL.detect\_points(image, obj\_name,num\_points)}]
  Identify instances of the named object by pointing to them with molmo.
  \begin{itemize}
    \item \textbf{Inputs}:
      \begin{itemize}
        \item \texttt{image}: PIL.Image
        \item \texttt{obj\_name}: string
        \item \texttt{num\_points}: Agent-selected maximum number of points
      \end{itemize}
    \item \textbf{Outputs}:
      \begin{itemize}
        \item \textbf{Raw value}: String serialization of a list of normalized point coordinates \((x,y) \in [0,1]^2\) 
        \item \textbf{Text}: Reports the object name, count, and the list of normalized points
        \item \textbf{Image}: Original image with red circular point markers at detected locations (white outlines)
        \item \textbf{Variables}: \texttt{\textit{<obj\_name>}\_detections} Stored variables: molmo\_obj\_name\_detections\_1, ...\_2, … (one per kept point; (x, y) in [0, 1]).
      \end{itemize}
  \end{itemize}

  \item[\texttt{UniDepthTool.pointcloud\_from\_mask(image, mask, output\_var)}]
    Extract a 3D point cloud from the region defined by a segmentation mask (intersected with valid-depth pixels).
    \begin{itemize}
      \item \textbf{Inputs}:
        \begin{itemize}
          \item \texttt{image}: PIL.Image or numpy array (RGB)
          \item \texttt{mask}: Boolean mask (H$\times$W), e.g.\ from Segmentation tool
          \item \texttt{output\_var}: Optional custom variable name suffix (e.g.\ \texttt{"coffee\_grinder\_points"})
        \end{itemize}
      \item \textbf{Outputs}:
        \begin{itemize}
          \item \textbf{Raw value}: \texttt{\{"point\_cloud": [[x,y,z], \ldots], "num\_points": N\}}
          \item \textbf{Text}: Reports number of extracted 3D points and variable reference
          \item \textbf{Image}: Original image with sampled points overlaid, colored by $Z$-depth via turbo colormap (color scale anchored to scene-wide $Z$
  range, 2\%–98\% percentile)
          \item \textbf{Variables}: \texttt{\$UniDepth\_\textit{output\_var}} or \texttt{\$UniDepth\_point\_cloud} — $N\times3$ list of $(x,y,z)$ in meters
  (camera frame)
        \end{itemize}
    \end{itemize}

  \item[\texttt{UniDepthTool.pointcloud\_full(image, output\_var)}]
    Extract the full-scene 3D point cloud for all pixels with valid metric depth.
    \begin{itemize}
      \item \textbf{Inputs}:
        \begin{itemize}
          \item \texttt{image}: PIL.Image or numpy array (RGB)
          \item \texttt{output\_var}: Optional custom variable name suffix
        \end{itemize}
      \item \textbf{Outputs}:
        \begin{itemize}
          \item \textbf{Raw value}: \texttt{\{"point\_cloud": [[x,y,z], \ldots], "num\_points": N\}}
          \item \textbf{Text}: Reports number of 3D points and image resolution ($H\times W$), with variable reference
          \item \textbf{Image}: None
          \item \textbf{Variables}: \texttt{\$UniDepth\_\textit{output\_var}} or \texttt{\$UniDepth\_point\_cloud} — $N\times3$ list of $(x,y,z)$ in meters
  (camera frame)
        \end{itemize}
    \end{itemize}

  \item[\texttt{UniDepthTool.pointcloud\_from\_point(image, x, y, output\_var)}]
    Retrieve the metric 3D position in the camera coordinate system for a single image location.
    \begin{itemize}
      \item \textbf{Inputs}:
        \begin{itemize}
          \item \texttt{image}: PIL.Image or numpy array (RGB)
          \item \texttt{x}: Normalized horizontal coordinate $x \in [0,1]$ (left$\to$right)
          \item \texttt{y}: Normalized vertical coordinate $y \in [0,1]$ (top$\to$bottom)
          \item \texttt{output\_var}: Optional variable name suffix (default stores as \texttt{\$UniDepth\_from\_point})
        \end{itemize}
      \item \textbf{Outputs}:
        \begin{itemize}
          \item \textbf{Raw value}: \texttt{\{"point\_3d": [x,y,z] | null, "point\_cloud": [[x,y,z]], "num\_points": 0|1, "pixel\_x": int, "pixel\_y": int,
  "normalized\_x": float, "normalized\_y": float\}}
          \item \textbf{Text}: 3D coordinates (m) and pixel index, or invalid-depth notice
          \item \textbf{Image}: Original image with the query point marked as a lime-colored dot
          \item \textbf{Variables}: \texttt{\$UniDepth\_\textit{output\_var}} or \texttt{\$UniDepth\_from\_point} — single-row list \texttt{[[x,y,z]]} in
  meters
        \end{itemize}
    \end{itemize}

  \item[\texttt{UniDepthTool.estimate\_focal\_length\_px(image)}]
    Estimate the camera focal length in pixels from a single RGB image using UniDepth's predicted intrinsics.
    \begin{itemize}
      \item \textbf{Inputs}:
        \begin{itemize}
          \item \texttt{image}: PIL.Image or numpy array (RGB)
        \end{itemize}
      \item \textbf{Outputs}:
        \begin{itemize}
          \item \textbf{Raw value}: \texttt{\{"focal\_length\_px": float\}}
          \item \textbf{Text}: Focal length value in pixels and variable reference
          \item \textbf{Image}: None
          \item \textbf{Variables}: \texttt{\$focal\_length\_px} — scalar float (mean of $f_x$ and $f_y$ from predicted $K$)
        \end{itemize}
    \end{itemize}

  \item[\texttt{DepthProTool.pointcloud\_from\_mask(image, mask, output\_var)}]
    Extract a 3D point cloud from the region defined by a segmentation mask (intersected with valid-depth pixels).
    \begin{itemize}
      \item \textbf{Inputs}:
        \begin{itemize}
          \item \texttt{image}: PIL.Image or numpy array (RGB)
          \item \texttt{mask}: Boolean mask (H$\times$W), e.g.\ from Segmentation tool
          \item \texttt{output\_var}: Optional custom variable name suffix (e.g.\ \texttt{"coffee\_grinder\_points"})
        \end{itemize}
      \item \textbf{Outputs}:
        \begin{itemize}
          \item \textbf{Raw value}: \texttt{\{"point\_cloud": [[x,y,z], \ldots], "num\_points": N\}}
          \item \textbf{Text}: Reports number of extracted 3D points and variable reference
          \item \textbf{Image}: Original image with sampled points overlaid, colored by $Z$-depth via turbo colormap (color scale anchored to scene-wide $Z$
  range, 2\%–98\% percentile)
          \item \textbf{Variables}: \texttt{\$DepthPro\_\textit{output\_var}} or \texttt{\$DepthPro\_point\_cloud} — $N\times3$ list of $(x,y,z)$ in meters
  (camera frame)
        \end{itemize}
    \end{itemize}

  \item[\texttt{DepthProTool.pointcloud\_full(image, output\_var)}]
    Extract the full-scene 3D point cloud for all pixels with valid metric depth.
    \begin{itemize}
      \item \textbf{Inputs}:
        \begin{itemize}
          \item \texttt{image}: PIL.Image or numpy array (RGB)
          \item \texttt{output\_var}: Optional custom variable name suffix
        \end{itemize}
      \item \textbf{Outputs}:
        \begin{itemize}
          \item \textbf{Raw value}: \texttt{\{"point\_cloud": [[x,y,z], \ldots], "num\_points": N\}}
          \item \textbf{Text}: Reports number of 3D points and image resolution ($H\times W$), with variable reference
          \item \textbf{Image}: None
          \item \textbf{Variables}: \texttt{\$DepthPro\_\textit{output\_var}} or \texttt{\$DepthPro\_point\_cloud} — $N\times3$ list of $(x,y,z)$ in meters
  (camera frame)
        \end{itemize}
    \end{itemize}

  \item[\texttt{DepthProTool.pointcloud\_from\_point(image, x, y, output\_var)}]
    Retrieve the metric 3D position in the camera coordinate system for a single image location.
    \begin{itemize}
      \item \textbf{Inputs}:
        \begin{itemize}
          \item \texttt{image}: PIL.Image or numpy array (RGB)
          \item \texttt{x}: Normalized horizontal coordinate $x \in [0,1]$ (left$\to$right)
          \item \texttt{y}: Normalized vertical coordinate $y \in [0,1]$ (top$\to$bottom)
          \item \texttt{output\_var}: Optional variable name suffix (default stores as \texttt{\$DepthPro\_from\_point})
        \end{itemize}
      \item \textbf{Outputs}:
        \begin{itemize}
          \item \textbf{Raw value}: \texttt{\{"point\_3d": [x,y,z] | null, "point\_cloud": [[x,y,z]], "num\_points": 0|1, "pixel\_x": int, "pixel\_y": int,
  "normalized\_x": float, "normalized\_y": float\}}
          \item \textbf{Text}: 3D coordinates (m) and pixel index, or invalid-depth notice
          \item \textbf{Image}: Original image with the query point marked as a lime-colored dot
          \item \textbf{Variables}: \texttt{\$DepthPro\_\textit{output\_var}} or \texttt{\$DepthPro\_from\_point} — single-row list \texttt{[[x,y,z]]} in
  meters
        \end{itemize}
    \end{itemize}

  \item[\texttt{DepthProTool.estimate\_focal\_length\_px(image)}]
    Estimate the camera focal length in pixels from a single RGB image using DepthPro's predicted intrinsics.
    \begin{itemize}
      \item \textbf{Inputs}:
        \begin{itemize}
          \item \texttt{image}: PIL.Image or numpy array (RGB)
        \end{itemize}
      \item \textbf{Outputs}:
        \begin{itemize}
          \item \textbf{Raw value}: \texttt{\{"focal\_length\_px": float\}}
          \item \textbf{Text}: Focal length value in pixels and variable reference
          \item \textbf{Image}: None
          \item \textbf{Variables}: \texttt{\$focal\_length\_px} — scalar float (mean of $f_x$ and $f_y$ from predicted $K$)
        \end{itemize}
    \end{itemize}

  \item[\texttt{MoGeTool.pointcloud\_from\_mask(image, mask, output\_var)}]
    Extract a 3D point cloud from the region defined by a segmentation mask (intersected with valid-depth pixels).
    \begin{itemize}
      \item \textbf{Inputs}:
        \begin{itemize}
          \item \texttt{image}: PIL.Image or numpy array (RGB)
          \item \texttt{mask}: Boolean mask (H$\times$W), e.g.\ from Segmentation tool
          \item \texttt{output\_var}: Optional custom variable name suffix (e.g.\ \texttt{"coffee\_grinder\_points"})
        \end{itemize}
      \item \textbf{Outputs}:
        \begin{itemize}
          \item \textbf{Raw value}: \texttt{\{"point\_cloud": [[x,y,z], \ldots], "num\_points": N\}}
          \item \textbf{Text}: Reports number of extracted 3D points and variable reference
          \item \textbf{Image}: Original image with sampled points overlaid, colored by $Z$-depth via turbo colormap (color scale anchored to scene-wide $Z$
  range, 2\%–98\% percentile)
          \item \textbf{Variables}: \texttt{\$MoGe\_\textit{output\_var}} or \texttt{\$MoGe\_point\_cloud} — $N\times3$ list of $(x,y,z)$ in meters
  (camera frame)
        \end{itemize}
    \end{itemize}

  \item[\texttt{MoGeTool.pointcloud\_full(image, output\_var)}]
    Extract the full-scene 3D point cloud for all pixels with valid metric depth.
    \begin{itemize}
      \item \textbf{Inputs}:
        \begin{itemize}
          \item \texttt{image}: PIL.Image or numpy array (RGB)
          \item \texttt{output\_var}: Optional custom variable name suffix
        \end{itemize}
      \item \textbf{Outputs}:
        \begin{itemize}
          \item \textbf{Raw value}: \texttt{\{"point\_cloud": [[x,y,z], \ldots], "num\_points": N\}}
          \item \textbf{Text}: Reports number of 3D points and image resolution ($H\times W$), with variable reference
          \item \textbf{Image}: None
          \item \textbf{Variables}: \texttt{\$MoGe\_\textit{output\_var}} or \texttt{\$MoGe\_point\_cloud} — $N\times3$ list of $(x,y,z)$ in meters
  (camera frame)
        \end{itemize}
    \end{itemize}

  \item[\texttt{MoGeTool.pointcloud\_from\_point(image, x, y, output\_var)}]
    Retrieve the metric 3D position in the camera coordinate system for a single image location.
    \begin{itemize}
      \item \textbf{Inputs}:
        \begin{itemize}
          \item \texttt{image}: PIL.Image or numpy array (RGB)
          \item \texttt{x}: Normalized horizontal coordinate $x \in [0,1]$ (left$\to$right)
          \item \texttt{y}: Normalized vertical coordinate $y \in [0,1]$ (top$\to$bottom)
          \item \texttt{output\_var}: Optional variable name suffix (default stores as \texttt{\$MoGe\_from\_point})
        \end{itemize}
      \item \textbf{Outputs}:
        \begin{itemize}
          \item \textbf{Raw value}: \texttt{\{"point\_3d": [x,y,z] | null, "point\_cloud": [[x,y,z]], "num\_points": 0|1, "pixel\_x": int, "pixel\_y": int,
  "normalized\_x": float, "normalized\_y": float\}}
          \item \textbf{Text}: 3D coordinates (m) and pixel index, or invalid-depth notice
          \item \textbf{Image}: Original image with the query point marked as a lime-colored dot
          \item \textbf{Variables}: \texttt{\$MoGe\_\textit{output\_var}} or \texttt{\$MoGe\_from\_point} — single-row list \texttt{[[x,y,z]]} in
  meters
        \end{itemize}
    \end{itemize}

  \item[\texttt{MoGeTool.estimate\_focal\_length\_px(image)}]
    Estimate the camera focal length in pixels from a single RGB image using MoGe's predicted intrinsics.
    \begin{itemize}
      \item \textbf{Inputs}:
        \begin{itemize}
          \item \texttt{image}: PIL.Image or numpy array (RGB)
        \end{itemize}
      \item \textbf{Outputs}:
        \begin{itemize}
          \item \textbf{Raw value}: \texttt{\{"focal\_length\_px": float\}}
          \item \textbf{Text}: Focal length value in pixels and variable reference
          \item \textbf{Image}: None
          \item \textbf{Variables}: \texttt{\$focal\_length\_px} — scalar float (mean of $f_x$ and $f_y$ from predicted $K$)
        \end{itemize}
    \end{itemize}

  \item[\texttt{GroundingDinoTool.detect\_2d\_bbox(image, obj\_name)}]
    Detect all instances of a named object in the image using Grounding DINO zero-shot open-vocabulary detection; results sorted by confidence score (highest
  first).
    \begin{itemize}
      \item \textbf{Inputs}:
        \begin{itemize}
          \item \texttt{image}: PIL.Image or numpy array (RGB)
          \item \texttt{obj\_name}: Target object phrase, modifiers included when the task disambiguates instances (e.g.\ \texttt{"right chair"} or
  \texttt{"orange bottle"} rather than a generic noun); words are internally hyphenated and appended with \texttt{" ."} as the detection caption
        \end{itemize}
      \item \textbf{Outputs}:
        \begin{itemize}
          \item \textbf{Raw value}: List of bounding boxes \texttt{[[x1,y1,x2,y2], \ldots]} in pixel integers (clamped to image bounds), sorted by
  GroundingDINO confidence score descending; empty list if none detected
          \item \textbf{Text}: Reports count, pixel-coordinate boxes, and per-box GroundingDINO scores; lists variable names if stored
          \item \textbf{Image}: Original image with all detected bounding boxes drawn as red rectangles
          \item \textbf{Variables}: \texttt{\$grounding\_dino\_\textit{obj\_name}\_bboxes\_1}, \texttt{\_2}, \ldots\ — one variable per detected box, each a
  list \texttt{[x1,y1,x2,y2]}
        \end{itemize}
    \end{itemize}

  \item[\texttt{Qwen3VLTool.detect\_2d\_bbox(image, obj\_name)}]
    Detect all instances of a named object in the image using Grounding DINO zero-shot open-vocabulary detection; results sorted by confidence score (highest
  first).
    \begin{itemize}
      \item \textbf{Inputs}:
        \begin{itemize}
          \item \texttt{image}: PIL.Image or numpy array (RGB)
          \item \texttt{obj\_name}: Target object phrase, modifiers included when the task disambiguates instances (e.g.\ \texttt{"right chair"} or
  \texttt{"orange bottle"} rather than a generic noun); words are internally hyphenated and appended with \texttt{" ."} as the detection caption
        \end{itemize}
      \item \textbf{Outputs}:
        \begin{itemize}
          \item \textbf{Raw value}: List of bounding boxes \texttt{[[x1,y1,x2,y2], \ldots]} in pixel integers (clamped to image bounds), sorted by
  Qwen3VL confidence score descending; empty list if none detected
          \item \textbf{Text}: Reports count, pixel-coordinate boxes, and per-box Qwen3VL scores; lists variable names if stored
          \item \textbf{Image}: Original image with all detected bounding boxes drawn as red rectangles
          \item \textbf{Variables}: \texttt{\$grounding\_dino\_\textit{obj\_name}\_bboxes\_1}, \texttt{\_2}, \ldots\ — one variable per detected box, each a
  list \texttt{[x1,y1,x2,y2]}
        \end{itemize}
    \end{itemize}

    \item[\texttt{WildDet3DTool.detect\_3d\_bbox\_from\\\_text(image, obj\_name)}]
    Detect all instances of an open-vocabulary category via WildDet3D text prompt and return full 3D detection results.
    \begin{itemize}
      \item \textbf{Inputs}:
        \begin{itemize}
          \item \texttt{image}: PIL.Image or numpy array (RGB)
          \item \texttt{obj\_name}: Category string (e.g.\ \texttt{"chair"}, \texttt{"red car"})
        \end{itemize}
      \item \textbf{Outputs}:
        \begin{itemize}
          \item \textbf{Raw value}: 2D boxes, 3D boxes, confidence scores, class IDs, and depth maps for all detections
          \item \textbf{Text}: Detection count, best score, selected OBB dimensions (m) and 3D centre (m)
          \item \textbf{Image}: Original image with projected 3D bounding boxes drawn
          \item \textbf{Variables}: OBB width, height, length (m), OBB size as a list, and OBB 3D centre — from the highest-score detection
        \end{itemize}
    \end{itemize}

  \item[\texttt{WildDet3DTool.detect\_3d\_bbox\_from\_bbox(image, obj\_name, bbox)}]
    Lift one 2D bounding box to 3D via WildDet3D geometric box prompt.
    \begin{itemize}
      \item \textbf{Inputs}:
        \begin{itemize}
          \item \texttt{image}: PIL.Image or numpy array (RGB)
          \item \texttt{obj\_name}: Label used only to name context variables
          \item \texttt{bbox}: Reference 2D box in pixel \texttt{xyxy}; compatible with QwenVL / GroundingDINO outputs
        \end{itemize}
      \item \textbf{Outputs}:
        \begin{itemize}
          \item \textbf{Raw value}: 2D boxes, 3D boxes, confidence scores, class IDs, and depth maps for all detections
          \item \textbf{Text}: Detection count, best score, selected OBB dimensions (m) and 3D centre (m)
          \item \textbf{Image}: Original image with projected 3D bounding boxes drawn
          \item \textbf{Variables}: OBB width, height, length (m), OBB size as a list, and OBB 3D centre — from the highest-score detection
        \end{itemize}
    \end{itemize}

  \item[\texttt{OVMono3DTool.detect\_3d\_bbox(image, obj\_name)}]
    Detect the 3D bounding box of a named object using OVMono3D open-vocabulary monocular 3D detector; returns the highest-scoring detection.
    \begin{itemize}
      \item \textbf{Inputs}:
        \begin{itemize}
          \item \texttt{image}: PIL.Image or numpy array (RGB)
          \item \texttt{obj\_name}: Category name to detect (e.g.\ \texttt{"chair"}, \texttt{"monitor"})
        \end{itemize}
      \item \textbf{Outputs}:
        \begin{itemize}
          \item \textbf{Raw value}: OBB size list, OBB 3D centre list, and raw detection records with scores
          \item \textbf{Text}: Detection count, OBB size [width, height, length] (m), 3D centre (m), and confidence scores
          \item \textbf{Image}: Original image with the detected 3D bounding box rendered as a cuboid mesh overlay
          \item \textbf{Variables}: OBB width, height, length (m), OBB size as a list, and OBB 3D centre — from the highest-score detection
        \end{itemize}
    \end{itemize}

    \item[\texttt{Qwen3VLTool.detect\_3d\_bbox\_from\_text(image, obj\_name)}]
    Detect all instances of an open-vocabulary category via Qwen3VL text prompt and return full 3D detection results.
    \begin{itemize}
      \item \textbf{Inputs}:
        \begin{itemize}
          \item \texttt{image}: PIL.Image or numpy array (RGB)
          \item \texttt{obj\_name}: Category string (e.g.\ \texttt{"chair"}, \texttt{"red car"})
        \end{itemize}
      \item \textbf{Outputs}:
        \begin{itemize}
          \item \textbf{Raw value}: 2D boxes, 3D boxes, confidence scores, class IDs, and depth maps for all detections
          \item \textbf{Text}: Detection count, best score, selected OBB dimensions (m) and 3D centre (m)
          \item \textbf{Image}: Original image with projected 3D bounding boxes drawn
          \item \textbf{Variables}: OBB width, height, length (m), OBB size as a list, and OBB 3D centre — from the highest-score detection
        \end{itemize}
    \end{itemize}
  \item[\texttt{BoundingBoxTool.detect\_3d\_bbox(point\_cloud, mask, object\_name, focal\_length\_px)}]
    Compute a minimal oriented 3D bounding box for the masked region of a point cloud using Open3D.
    \begin{itemize}
      \item \textbf{Inputs}:
        \begin{itemize}
          \item \texttt{point\_cloud}: Scene point cloud, $N\times3$ array in meters.
          \item \texttt{mask}: Boolean mask (H$\times$W) segmenting the target object.
          \item \texttt{object\_name}: Label used to name output variables (e.g.\ \texttt{"chair"})
          \item \texttt{focal\_length\_px}: Camera focal length in pixels ($f_x = f_y$ assumed); pass \texttt{\$focal\_length\_px} from the depth tool
        \end{itemize}
      \item \textbf{Outputs}:
        \begin{itemize}
          \item \textbf{Raw value}: OBB size, OBB 3D centre, 8 corner points in 3D (OpenCV frame) and normalized 2D, edge list, and raw extent
          \item \textbf{Text}: Number of points used, OBB size [width, height, length] (m), and 3D centre (m)
          \item \textbf{Image}: None
          \item \textbf{Variables}: OBB width, height, length (m), OBB size as a list, and OBB 3D centre
        \end{itemize}
    \end{itemize}
\end{description}
\FloatBarrier

%% file: section/case_study.tex
\section{Case Study}

\begin{figure}[h]
\centering

\begin{subfigure}{\columnwidth}
    \centering
    \includegraphics[width=\columnwidth]{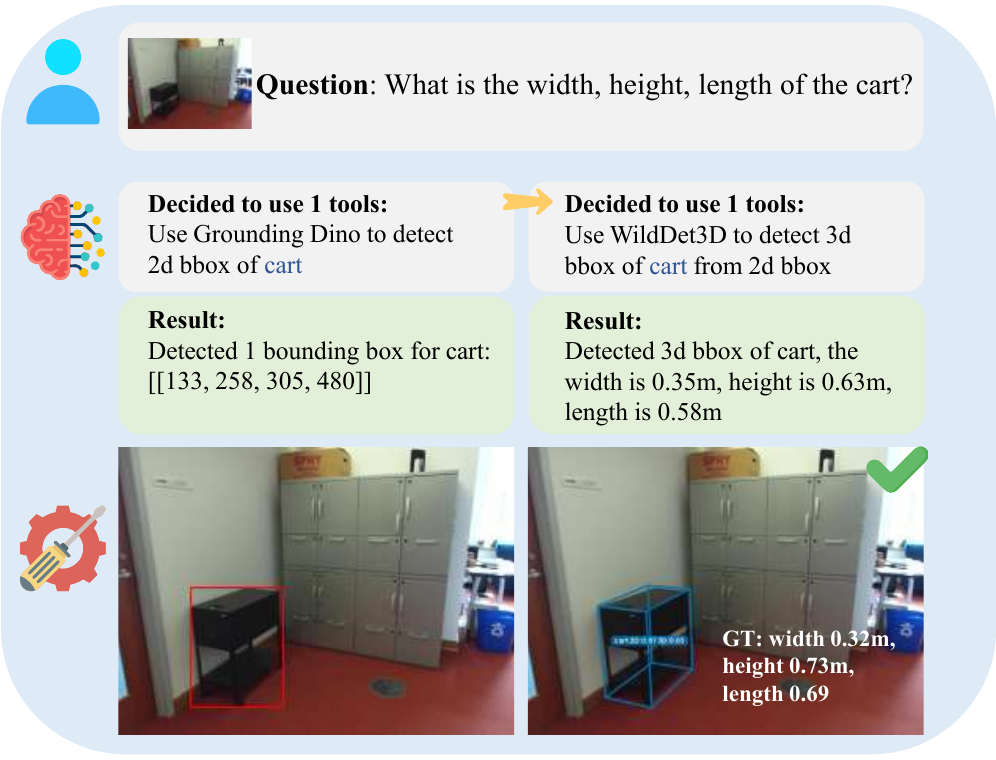}
    \caption{Indoor object size estimation.}
    \label{fig:case_study_5}
\end{subfigure}

\vspace{1ex}

\begin{subfigure}{\columnwidth}
    \centering
    \includegraphics[width=\columnwidth]{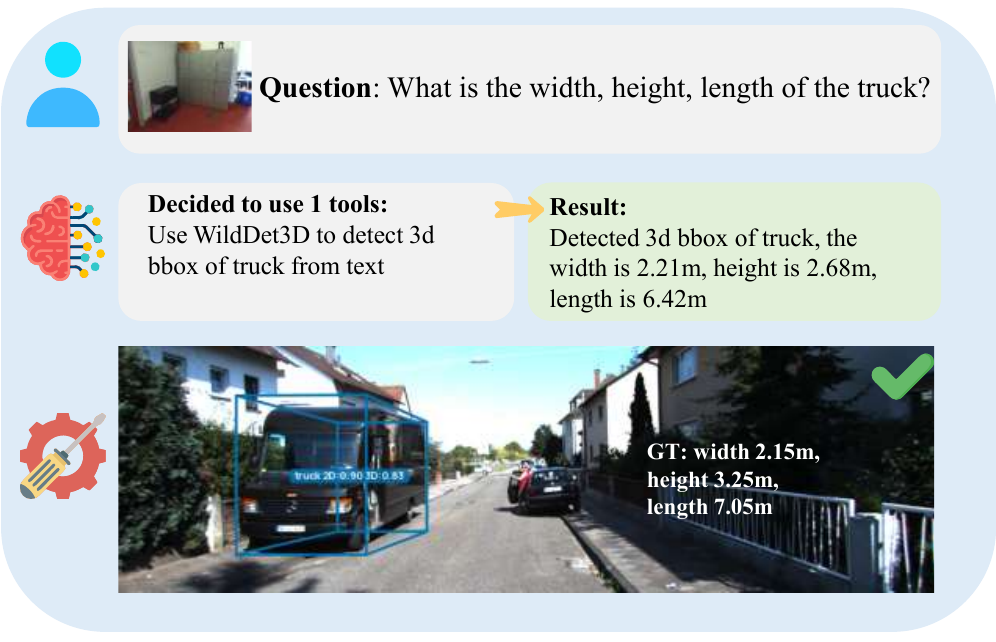}
    \caption{Outdoor object size estimation.}
    \label{fig:case_study_6}
\end{subfigure}

\caption{Examples of \ModelNameL~solving object size estimation problems in indoor and outdoor scenes.}
\vspace{-2ex}
\label{fig:case_study_size}
\end{figure}

\begin{figure*}[t]
\centering

\begin{subfigure}{\textwidth}
    \includegraphics[width=\textwidth]{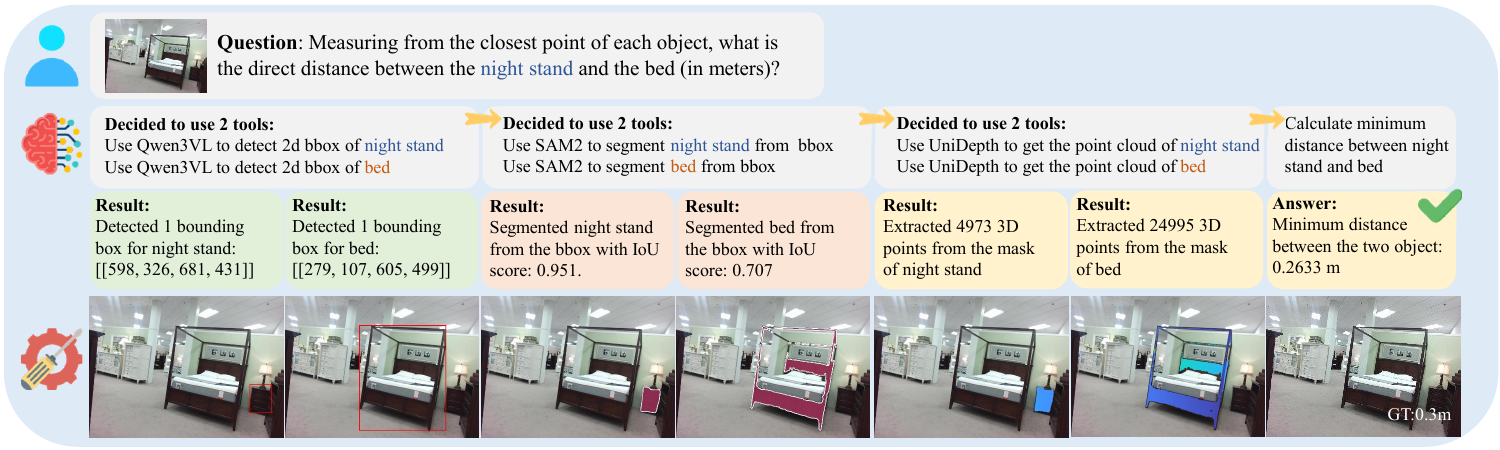}
    \caption{Indoor object-object distance estimation.}
\end{subfigure}

\vspace{1ex}

\begin{subfigure}{\textwidth}
    \includegraphics[width=\textwidth]{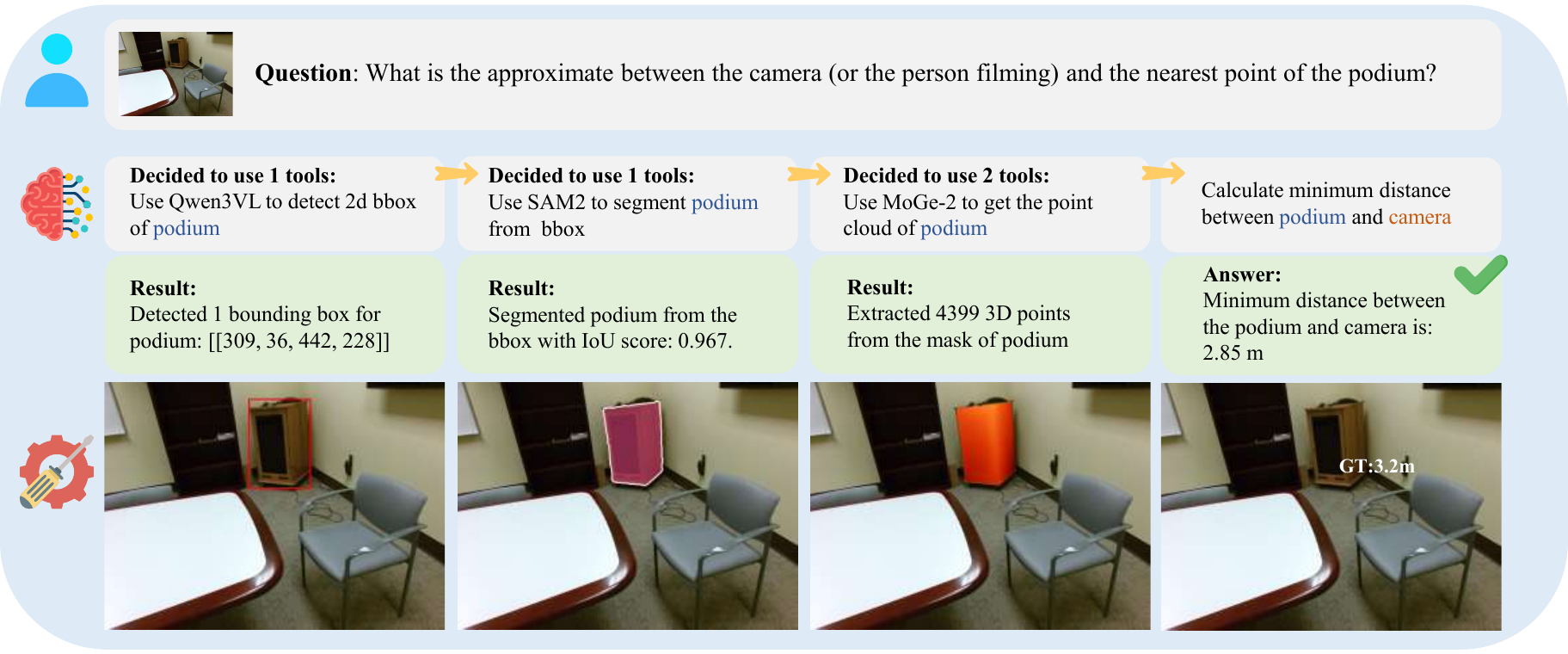}
    \caption{Indoor object-camera distance estimation.}
\end{subfigure}

\vspace{1ex}

\begin{subfigure}{\textwidth}
    \includegraphics[width=\textwidth]{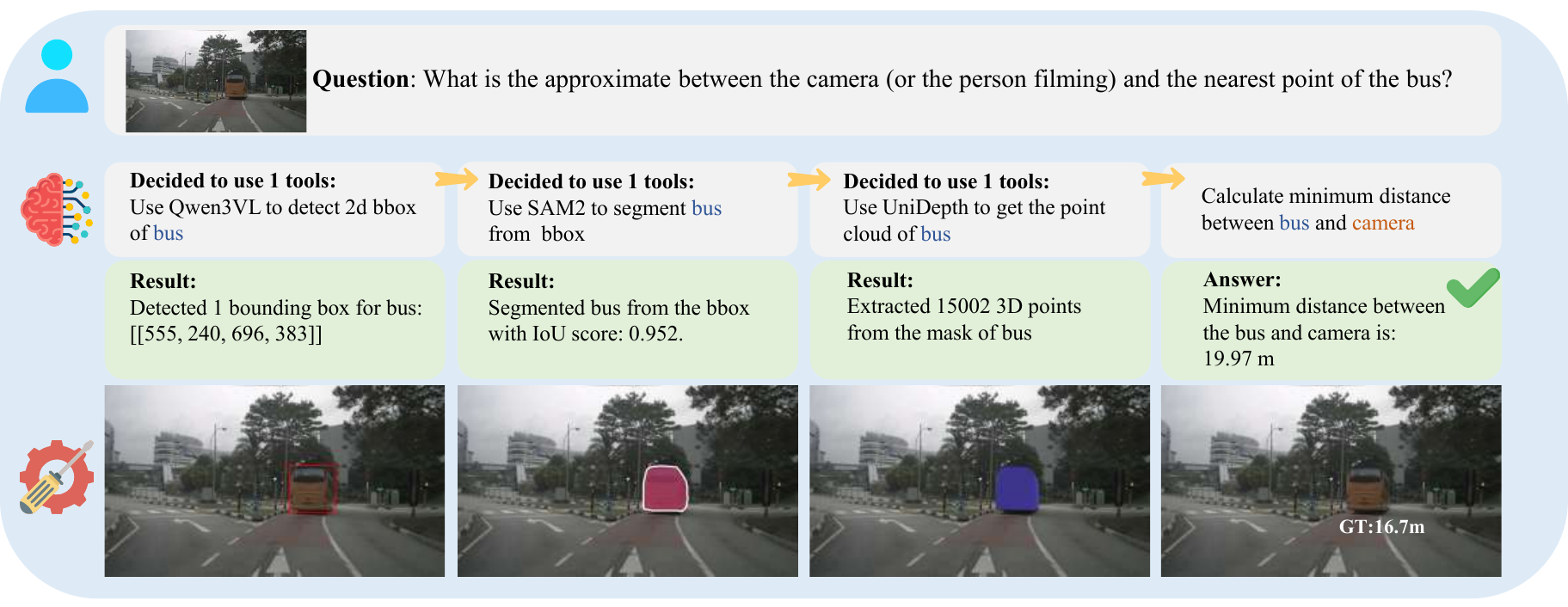}
    \caption{Outdoor object-camera distance estimation.}
\end{subfigure}

\caption{Case studies of \ModelNameL~on quantitative spatial reasoning tasks.}
\label{fig:case_study_all}

\end{figure*}

%% file: section/prompt.tex
\section{Prompt}~\label{app:prompt}

\definecolor{promptbg}{RGB}{248,249,250}
\definecolor{promptborder}{RGB}{218,220,224}
\definecolor{prompttitle}{RGB}{26,13,171}

\tcbset{
    promptbox/.style={
        colback=promptbg,
        colframe=promptborder,
        boxrule=1pt,
        arc=4pt,
        left=8pt,
        right=8pt,
        top=8pt,
        bottom=8pt,        fonttitle=\bfseries\color{prompttitle}
    }
}

\subsection{Prompt for Evaluation}
We use GPT-4o to evaluate the results. For quantitative queries,GPT-4o extracts
numerical values from the responses, and we calculate the success rate using a predefined threshold. For qualitative questions, GPT-4o scores the model’s answers between 0 and 1. 
\begin{tcolorbox}[promptbox, title=Prompt for extract scalar result]
\begin{Verbatim}[fontsize=\scriptsize, breaklines=true]
You are a precise measurement extraction system.

You must:
- Extract numerical values from model answers
- Evaluate arithmetic expressions
- Handle approximate values (about)
- For ranges, use midpoint
- Convert all outputs to SI units (metres)

Unit conversion:
1 inch = 0.0254 m
1 foot = 0.3048 m
1 cm = 0.01 m
1 mm = 0.001 m
1 m = 1 m

Output must be:
- A single number in metres OR null
- No explanation, no units

Task: extract a single scalar measurement in metres.

Question:
{question}

Answer:
{raw}

Return ONLY one number or null.
\end{Verbatim}
\end{tcolorbox}

\begin{tcolorbox}[promptbox, title=Prompt for extract list result]
\begin{Verbatim}[fontsize=\scriptsize, breaklines=true]
You are a precise measurement extraction system.

You must:
- Extract numerical values from model answers
- Evaluate arithmetic expressions
- Handle approximate values (about)
- For ranges, use midpoint
- Convert all outputs to SI units (metres)

Unit conversion:
1 inch = 0.0254 m
1 foot = 0.3048 m
1 cm = 0.01 m
1 mm = 0.001 m
1 m = 1 m

Output must be:
- A list of numbers in metres OR null
- No explanation, no units

Task: extract {n} numerical measurements in metres.

Order matters and should follow the semantics of the question.

Question:
{question}

Answer:
{raw}

Return ONLY {n} comma-separated numbers or null.
\end{Verbatim}
\end{tcolorbox}

\begin{tcolorbox}[promptbox, title=Prompt for qualitative problem]
\begin{Verbatim}[fontsize=\scriptsize, breaklines=true]
You should help me evaluate whether a model response matches the correct answer given a question.

You must output a single integer:
- 1 means the response is semantically correct and matches the answer.
- 0 means the response is incorrect or inconsistent with the answer.

Your output must be in valid JSON format with the key "your_mark".

Example:
Question: Between the wooden table and the black chair, which one is taller?
Answer: The wooden table is taller.
Response: The chair.
Output: {"your_mark": 0}

Your Turn:
Question: [Question]
Answer: [Answer]
Response: [Response]

Output:
\end{Verbatim}
\end{tcolorbox}

\subsection{Prompt for perceiver Model}
The Instruction prompt guides the perceiver to classify the input image and query into predefined scene and task categories for memory routing.

\begin{tcolorbox}[promptbox, title=Prompt for Perceiver Model]
\begin{Verbatim}[fontsize=\scriptsize, breaklines=true]
You are a spatial-question classifier.

Given an image and a spatial measurement question, output JSON only with no markdown and no explanation.

Question: {question}

Classify the following from the image and question:

1. scene:
Determine whether the image is indoors or outdoors.
Return one of: "Indoor", "Outdoor", "Unknown".

2. question_type:
Identify the type of spatial measurement:
"OO" = distance between two objects
"OC" = distance from an object to the camera
"OS" = size or dimension of an object (width, height, length, etc.)
Return one of: "OO", "OC", "OS", "Unknown".

Return format (JSON only):
{"scene": "...", "question_type": "..."}

\end{Verbatim}
\end{tcolorbox}

\subsection{Prompt for summarizer Model}
The instruction prompt guides the summarizer to extract a compact tool trajectory by retaining only the tool sequence from the full execution trace.

\begin{tcolorbox}[promptbox, title=Prompt for Summarizer Model, width=\linewidth]
\begin{Verbatim}[fontsize=\scriptsize, breaklines=true]
You are an expert in multi-modal tool-using agents (vision, geometry, segmentation, 3D, etc.).

You are given a user query and multiple trajectories for the same task. Text describes tool calls, and images may be provided in execution order with a figure index. Use both text and images to evaluate different runs.

Tool naming (critical):
Both the registry name and the tool category matter. Multiple tools may belong to the same category (e.g., 2D detection, segmentation, depth estimation). You must always refer to the exact registry name and not collapse different tools into a single abstract category.

When writing insights, use the format:
registry_name (category)

For example:
grounding_dino (2D bounding box / open-vocabulary grounding)
sam3 (instance segmentation)
moge (depth / point cloud & intrinsics)

Always distinguish tools by their exact registry names even if they share the same category.

Visual evidence:
Images are provided in execution order, where each figure corresponds to the output of a specific tool call. Do not misalign images and tools. Use images to verify and compare tool outputs such as masks, depth maps, or point clouds.

Task:
Compare all trajectories and identify the quality_rank=1 trajectory (best one). Determine whether it is successful or failed.

Part 1 — Current Trajectory Summary:
- Status: success or fail
- Description: one sentence summary of what happened
- Pipeline: ordered tool chain using exact registry names

Part 2 — Optimized Trajectory:
If the best trajectory succeeded:
- Keep the same pipeline
- Status: same
- Rationale: it already succeeds

If it failed:
- Propose an improved pipeline
- Only substitute tools within the same functional category
- Keep correct steps unchanged
- Explain briefly why in one sentence

Allowed substitutions include:
- 2D detection: grounding_dino, qwenvl
- point selection: molmo, qwenvl
- segmentation: sam2, sam3
- depth / point cloud: depth_pro, moge, unidepth
- 3D detection: pointcloud_to_bbox, qwenvl, wilddet3d, ovmono3d

Output format (strict):

# Current Trajectory
Status: success or fail
Description: <one sentence>
Pipeline: tool1 → tool2 → tool3

<<<OPTIMIZED_TRAJECTORY>>>

# Optimized Trajectory
Status: same or improved
Rationale: <one sentence>
Pipeline: tool1 → tool2 → tool3

\end{Verbatim}
\end{tcolorbox}